%% file: main.tex
\pdfoutput=1

\documentclass[11pt]{article}

\usepackage[]{ACL2023}

\usepackage{times}
\usepackage{latexsym}
\usepackage{multirow}
\usepackage{graphicx}
\usepackage{subcaption}
\usepackage{booktabs}
\usepackage{hyperref}
\usepackage{url}
\usepackage{comment}
\usepackage{xspace}
\usepackage{wrapfig}

\newcommand{\eat}[1]{} 

\newcommand{\answerTODO}[1][]{\textcolor{red}{\bf [TODO]}}

\usepackage[T1]{fontenc}

\usepackage[utf8]{inputenc}

\usepackage{microtype}

\usepackage{inconsolata}
\usepackage{enumitem}
\usepackage{algorithm}
\usepackage[noend]{algpseudocode}

\title{AutoMoE: Heterogeneous Mixture-of-Experts with Adaptive Computation for Efficient Neural Machine Translation}

\author{Ganesh Jawahar\thanks{\ \ Correspondence to \{ganeshjwhr@gmail.com,\\ subhabrata.mukherjee@microsoft.com\}.}$^{\clubsuit}$, Subhabrata Mukherjee$^\spadesuit$ \\ \textbf{Xiaodong Liu$^\spadesuit$, Young Jin Kim$^\spadesuit$, Muhammad Abdul-Mageed$^{\clubsuit}$$^\diamondsuit$, Laks V.S. Lakshmanan$^{\clubsuit}$}\\ \textbf{Ahmed Hassan Awadallah$^{\spadesuit}$, Sebastien Bubeck$^{\spadesuit}$, Jianfeng Gao$^{\spadesuit}$} \\\\ \normalsize $^\clubsuit$University of British Columbia, $^\spadesuit$Microsoft Research, $^\diamondsuit$MBZUAI  \\
 }

\newcommand{\sysname}{\texttt{AutoMoE}}

\begin{document}
\maketitle

\input{abstract}
\input{intro}

\input{relatedwork}
\input{proposal}

\input{experiments}

\input{results}
\input{conclusion}

\input{limitations.tex}
\input{ack}

\bibliography{anthology,main}
\bibliographystyle{acl_natbib}

\appendix
\input{appendix}

\end{document}

%% file: abstract.tex
\begin{abstract}

Mixture-of-Expert (MoE) models have obtained state-of-the-art performance in Neural Machine Translation (NMT) tasks. Existing works in MoE mostly consider a homogeneous design where the same number of experts of the same size are placed uniformly throughout the network. Furthermore, existing MoE works do not consider computational constraints (e.g., FLOPs, latency) to guide their design. To this end, we develop {\sysname} -- a framework for designing heterogeneous MoE's under computational constraints. 
{\sysname} leverages Neural Architecture Search (NAS) to obtain efficient sparse MoE sub-transformers with $4\times$ inference speedup (CPU) and FLOPs reduction over manually designed Transformers, 
with parity in BLEU score over dense Transformer and within $1$ BLEU point of MoE SwitchTransformer, on aggregate over benchmark datasets for NMT.
Heterogeneous search space with dense and sparsely activated Transformer modules (e.g., {\em how many experts? where to place them? what should be their sizes?}) allows for adaptive compute -- where different amounts of computations are used for different tokens in the input. Adaptivity comes naturally from routing decisions which send tokens to experts of different sizes. 
{{\sysname} code, data, and trained models are available at \url{https://aka.ms/AutoMoE}.}

\end{abstract}

%% file: intro.tex
\section{Introduction}
\label{sec:introduction}

\begin{figure*}
\centering
\includegraphics[width=5in,height=2.8in]{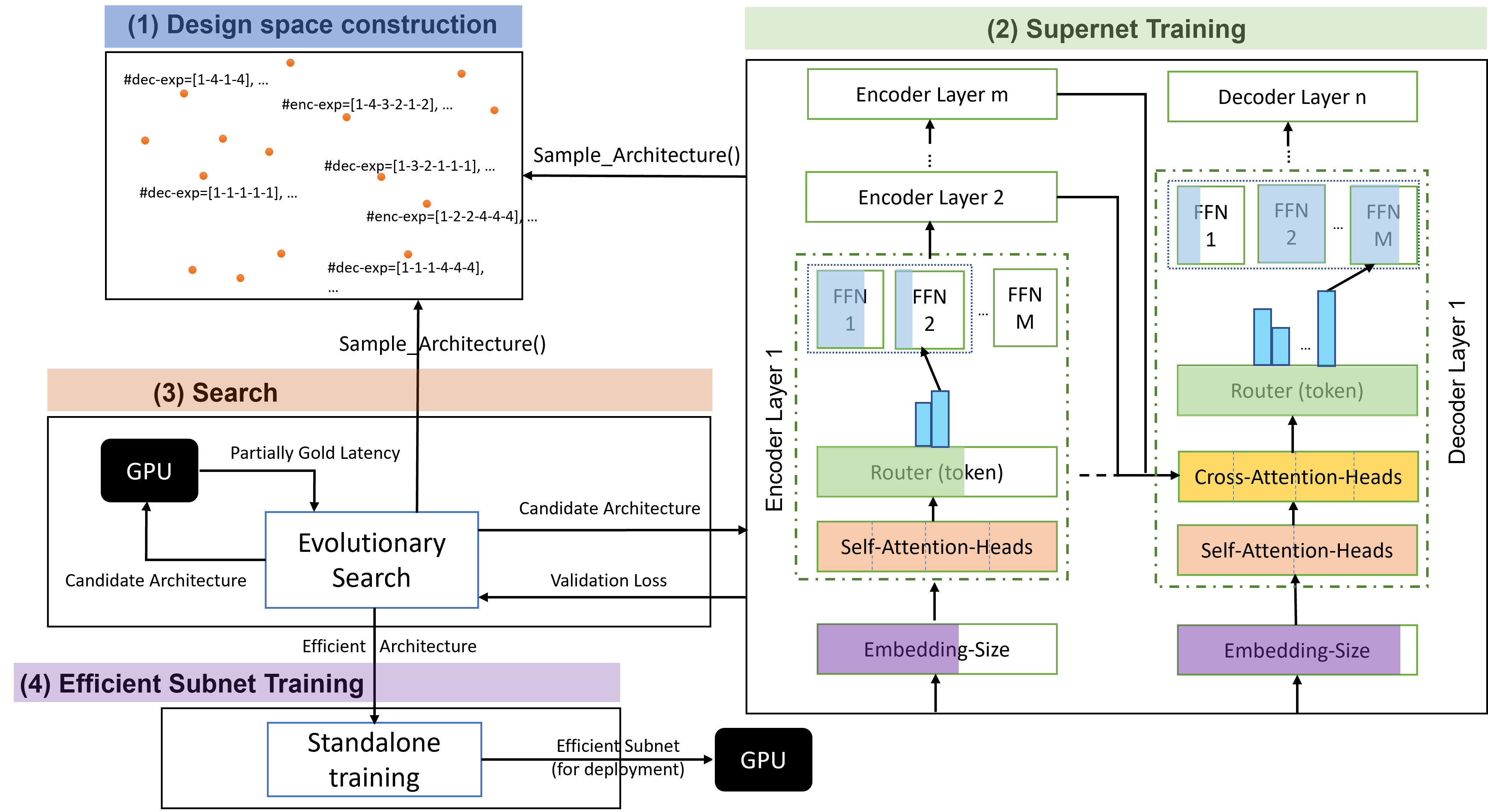}
\caption{{\sysname} Framework. (1) Heterogeneous MoE with variable dimensions for dense Transformer blocks and sparsely activated expert modules. (2) Supernet training by sampling subnetworks from search space and training them by sharing common weights with Supernet. (3) Evolutionary search to find efficient architectures by (a) sampling MoE subnetworks from the search space; (b) using latency measured in the target device; and (c) performance estimation from Supernet as feedback for iterative optimization via crossover and mutation. (4) Efficient MoE subnetwork(s) from evolutionary search is trained on downstream task. }
\label{fig:framework}
\vspace*{-0.5em}
\end{figure*}

Sparsely activated models like the Mixture-of-Experts (MoE)~\citep{switchtransformers} perform conditional computation in which only a subset of the weights of the network are activated per input. Selective compute allows us to design neural networks with a large number of model parameters,  without significant increase in the computational cost. With increased capacity, these sparse models have demonstrated state-of-the-art performance in natural language tasks such as neural machine translation (NMT)~\citep{kim_etal_21,kudugunta_etal_21,stochasticexperts}.

MoE architectures require several design choices: {\em (a) Expert placement:} Identifying Transformer layers for introducing expert sub-networks. {\em (b) Number of experts:} How many experts to place in different layers? {\em (c) Expert FFN size}: What should be the feedforward network (FFN) size for each expert? Given the large search space of potential architectures and the exorbitant computational cost of training and evaluating them, existing approaches manually design MoE architectures from a highly-restricted homogeneous space. For instance, they  use the same number of experts of the same capacity in different layers and make ad-hoc decisions like introducing experts in every other layer~\citep{switchtransformers,kim_etal_21,stochasticexperts, glam,artetxe_arxiv21} or every four layers~\citep{stmoe}. 

While these MoE's support conditional computation, homogeneity (specifically, fixed-size experts) results in the same amount (albeit different subsets) of weights to be applied to each input. We hypothesize that this is not an optimal solution and that we can reduce the number of experts (in some layers) to reduce communication cost, and the size (of some experts) to reduce computation cost resulting in reduction in model size, FLOPs and latency without much quality degradation.

This naturally extends MoEs to be adaptive compute models (similar to work on early exit~\cite{calm}) where different amounts of computations are used for different inputs. The adaptivity comes naturally from the routing decisions which would send tokens to experts of different sizes.

The above observations are depicted in Table~\ref{tab:heuristics_vs_search_motivation}, which shows demonstrative examples of manually designed MoE's vs. those designed by our {\sysname} framework. We compare these architectures against various computational metrics (e.g., latency, FLOPs, active MoE parameters), architectural configurations and task performance. For the most efficient configuration (last row in the table), {\sysname} reduces the number of decoder layers, compensating for the capacity with increased experts in the bottom layer, and places most of the experts in the encoder. Overall \sysname\ introduces the following components and contributions:

\begin{itemize}[leftmargin=*]
    \item \textit{Heterogeneous design with adaptive computation} for MoEs with variable number, size and placement of experts in both encoders and decoders. 
    \item Extends \textit{Supernet training} and evolutionary search from prior work on dense Transformers to new search space of sparse MoE's. This combines all possible MoE sub-architectures in a single graph; jointly training them via weight-sharing; and searching for optimal one with best possible performance on a downstream task satisfying a user-specified computational constraint.
    \item Experiments on NMT benchmarks demonstrate {\sysname}-designed MoE's to obtain $4\times$ inference speedup on CPU and equal FLOPs reduction over manually designed Transformers, with parity in BLEU with dense Transformer and within $1$ BLEU point of MoE SwitchTransformer. Further, it outperforms NAS methods in the dense search space (e.g., $1.3\times$ and $2.4\times$ FLOPs reduction and inference speedup over HAT~\citep{hat} and Evolved Transformer~\cite{evotrans}). 
    \end{itemize}

\begin{table*}[htb]
\scriptsize
\begin{center}
\begin{tabular}{p{1.5in}p{0.5in}p{0.5in}p{0.3in}p{0.35in}p{0.7in}p{0.5in}} \toprule
\textbf{Machine Translation} & \multicolumn{2}{c}{\textbf{\#Experts in each layer}} & \textbf{Accuracy} & \multicolumn{3}{c}{\textbf{Computational Footprint}} \\ \midrule
\textbf{Design Approach} & \textbf{Encoder} & \textbf{Decoder} & \textbf{BLEU} & \textbf{Latency} & \textbf{\# Active Params} & \textbf{FLOPs (G)} \\ \midrule
Manually designed (every layer) & 4-4-4-4-4-4 & 4-4-4-4-4-4 & 27.87 & 861ms  & 56M & 3.4 \\
Manually designed (every other layer) & 1-4-1-4-1-4 & 1-4-1-4-1-4 & 28.48 & 794ms  & 56M & 3.4 \\
\sysname & 1-1-4-4-4-1 & 4-1-1-1 & 28.15 & 585ms  & 46M & 2.9 \\ 
\bottomrule

\end{tabular}
\caption{Manual vs. {\sysname} designed MoE for illustration with $6$-layer encoder-decoder Transformer. Detailed results in Table~\ref{tab:nasmoe_vs_baseline}. We report computational metrics (measured on Intel Xeon CPU) and BLEU score of MoE's on WMT'14 En-De MT task. Number of experts per layer are separated by hyphen (-) for encoder and decoder.} 
\label{tab:heuristics_vs_search_motivation}
\end{center}
\end{table*}
%

%% file: relatedwork.tex
\section{Background}
\label{sec:background}


\textbf{Mixture-of-Experts:} 
MoE's have a rich literature in machine learning dating back to the 
early 90s~\citep{20yearsmoe}. 
They have received significant attention with works such as \cite{shazeer2017}, Switch Transformers~\citep{switchtransformers}, GShard~\citep{gshard}, BASE~\citep{base}, Hash~\citep{hash}, GLaM~\citep{glam}, Stochastic Experts~\citep{stochasticexperts}, Gating Dropout~\citep{gating_dropout} and  ST-MoE~\citep{stmoe}. Some crucial differences in these works include choice of expert routing function, expert placement technique, stability/performance enhancement techniques and nature of the task (pre-training vs. fine-tuning). Some challenges in building sparse expert models include: (i) lack of diversity in expert design (expert layer selection, number of experts, expert size, etc.), (ii) training instability, (iii) poor out-of-distribution generalization, (iv) cross-task adaptation of pre-trained models, (v) communication bottleneck, (vi) high memory and (vii) expert load balancing issue, to name a few. A comprehensive review of recent sparse expert models can be found at \cite{fedus_survey_arxiv22}.

\noindent \textbf{MoE design:} Most works in MoE rely on ad-hoc manual choices for expert placement, number of experts and their sizes. Existing approaches mostly use manual design, where they add experts on (i) alternate layers~\citep{switchtransformers,kim_etal_21,stochasticexperts,glam,artetxe_arxiv21}, (ii) every four layers~\citep{stmoe}, or (iii) final few layers~\cite{deepspeed_moe}. While these MoE's support conditional computation, they generally do not support adaptive compute since same number of expert parameters apply to every input, largely given by their homogeneous design (e.g., all experts of same size). 
Further, MoE design is generally agnostic to 
computational constraints (e.g., latency, memory) of the hardware in which the MoE model has to be deployed.


\noindent \textbf{Neural Architecture Search (NAS):} Given a search space of architectures and efficiency constraints (e.g., model size, latency), NAS typically aims to identify the optimal architecture that maximizes the task performance, while satisfying the efficiency constraints. 
NAS has been recently used for natural language understanding tasks to build efficient BERT~\citep{bert} and GPT~\citep{gpt3} based pre-trained language models~\citep{nasbert,autotinybert,autodistill,magic,autobert_zero,efficient_bert,primer,litetransformersearch} as well as for machine translation tasks~\citep{evotrans,hat}. Hardware aware transformers (HAT)~\citep{hat} is a state-of-the-art NAS framework with dense Transformers for MT that uses hardware latency as feedback for optimization.

However, all of the above NAS works consider a search space with densely activated Transformers and non-MoE architectures, They primarily search over typical Transformer architectural hyper-parameters like number of layers, attention heads and hidden size. In contrast, we propose the first NAS framework that searches for efficient sparsely activated Mixture-of-Expert modules in Transformers.
Our heterogeneous {\sysname} framework addresses some longstanding design choices for MoE's like {\em how many experts? which layers to place them? what should be their sizes?} and so on.


%% file: proposal.tex
\section{Designing Heterogeneous Mixture-of-Experts}
\label{sec:prop_framework}

We now present the components of \sysname~framework (illustrated in Figure~\ref{fig:framework}) for designing efficient MoE's under computational constraints.

\subsection{Heterogeneous MoE Search Space}
\label{sec:search_space}

\begin{table*}
\scriptsize
\begin{center}
\begin{tabular}{p{2in}p{1in}p{1in}} \toprule
\textbf{Attributes} & \textbf{{\sysname}} & \textbf{Transformer Base / Big} \\ \midrule
 Encoder-Embedding-Size & \{512, 640\} & 512 / 1024 \\
 Decoder-Embedding-Size & \{512, 640\} & 512 / 1024\\
 \#Encoder-Layers & \{6\} & 6 \\
 \#Decoder-Layers & \{1, 2, 3, 4, 5, 6\} & 6\\
 Encoder-QKV-Dim & \{512\} & 512 / 1024\\
 Decoder-QKV-Dim & \{512\} & 512 / 1024\\
 \#Encoder-Self-Att-Heads (PL) & \{4, 8\} & 8 / 16  \\
 \#Decoder-Self-Att-Heads (PL) & \{4, 8\} & 8 / 16\\
 \#Decoder-Cross-Att-Heads (PL) & \{4, 8\} & 8 / 16\\
 \#Decoder-Arbitrary-Att (PL) &\{-1, 1, 2\} & -1 \\\midrule
 Encoder-FFN-Intermediate-Size (PL, PE) & \{1024, 2048, 3072\} & 2048 / 4096 \\
 Decoder-FFN-Intermediate-Size (PL, PE) & \{1024, 2048, 3072\}  & 2048 / 4096\\ 
 \#Encoder-Experts (PL) & \{1, 2, $\cdots$ M\} & - \\
\#Decoder-Experts (PL) &  \{1, 2, $\cdots$ M\} & - \\
 \bottomrule
\end{tabular}
\vspace{-1em}
\caption{Search space of {\sysname} compared to manually configured Transformer Base / Big. `PL' and `PE' refer to per layer and per expert search dimensions. Decoder arbitrary attn. searches last $k$ encoder layers to attend for each decoder layer. FFN size varies across layers and experts. $M$ denotes maximum experts per layer.}
\label{tab:design_space}
\end{center}
\vspace*{-2em}
\end{table*}

Existing MoE approaches restrict their design space by  considering uniform distribution of size and number of experts placed in different Transformer layers. For instance, the standard MoE design~\citep{switchtransformers} for an $L$-layer Transformer with $M$ experts placed in alternate layers have only two possible configurations viz., {\small $\{1$-$M$-$1$-$\cdots$ $\}, \{M$-$1$-$M$- $\cdots \}$}. (a) Our design space allows {\em variable number of experts} in each layer resulting in ${M}^L$ possible configurations. (b) Furthermore, our design space also allows {\em variable expert size}, e.g., by modulating the width of the feedforward (FFN) subnetworks for different experts. Considering $N$ possible FFN dimensions for each expert results in ${N^M}^L$ possible configurations for designing the expert space. (c) Finally, given the autoregressive nature of tasks like neural machine translation, the inference cost is dominated by the decoder~\citep{kasai2021deep}. For instance, for token-based MoE, decoders take $200 \times$ the time per step compared to encoders at peak throughput~\citep{kudugunta_etal_21}. Therefore, we further consider {\em variable number of decoder layers} along with the above choices for expert placement and expert capacity. \textit{\textbf{To the best of our knowledge, our work is the first to study such a flexible and exhaustive design space for MoE architectures}}.

In addition to heterogeneous experts, we allow flexible design for non-expert Transformer modules like the number of attention heads, hidden size and intermediate feedforward dimensions. This heterogeneous design of non-MoE, i.e.,  dense Transformer modules, has been explored in prior works such as HAT~\citep{hat} for generation tasks like NMT, and AutoDistil~\citep{autodistill} for understanding tasks like those in the GLUE benchmark~\cite{wang-etal-2018-glue}. Table~\ref{tab:design_space} shows our search space. We demonstrate our heterogeneous MoE search to perform better than both manual and NAS-searched architectures in the dense space.

\subsection{Supernet Training for MoE}
\label{sec:supernet training}
\begin{figure*}[t!]
    \centering
    \begin{subfigure}[b]{0.25\textwidth}
        \centering
        \includegraphics[height=1.4in, width=1.5in]{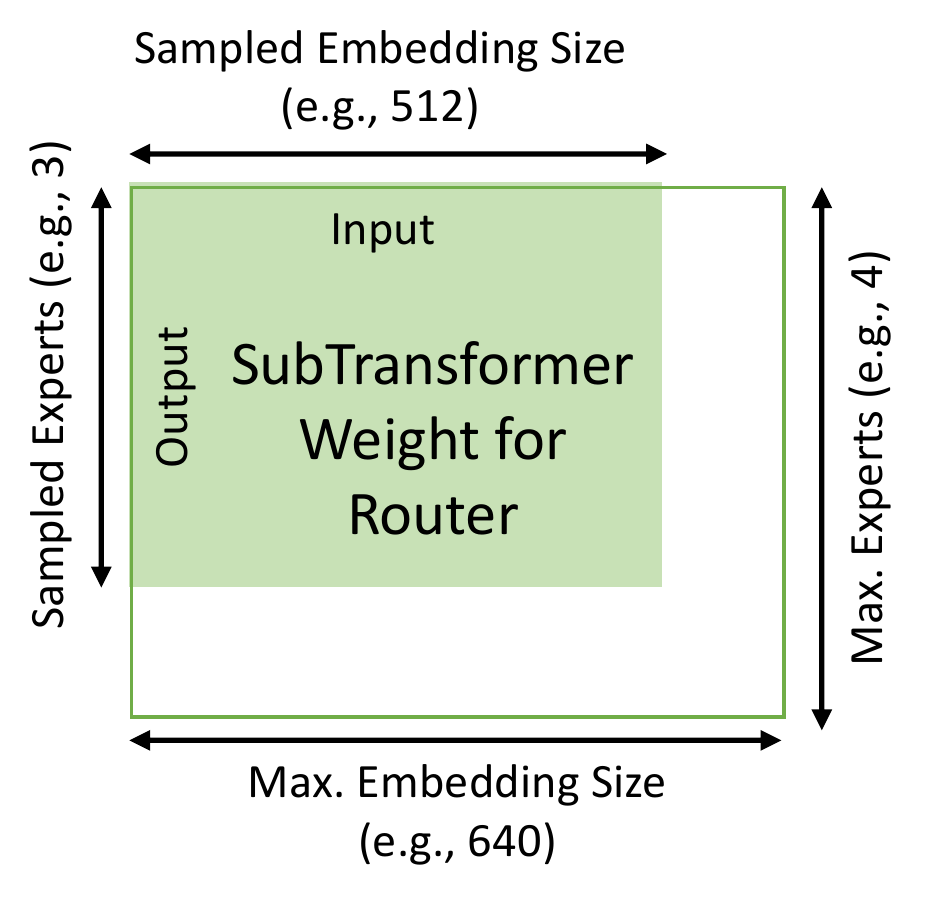}
        \caption{Router}
    \end{subfigure}%
    ~ 
    \begin{subfigure}[t]{0.8\textwidth}
        \centering
        \includegraphics[height=2.5in, width=3.6in]{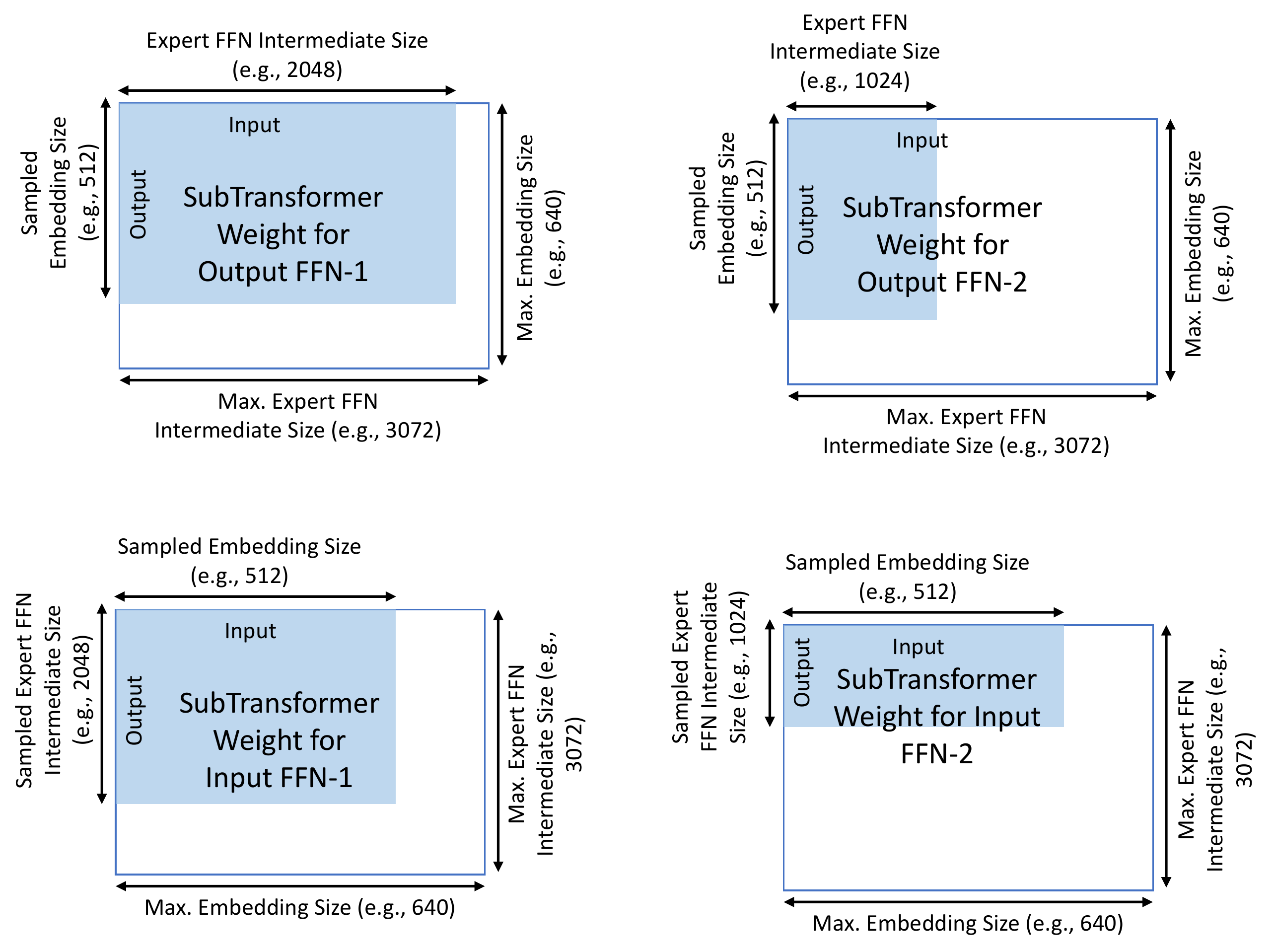}
        \caption{Experts (e.g., 2 FFN experts)}
    \end{subfigure}
    \vspace{-0.7em}
    \caption{Weight sharing in the MoE Supernet for sparsely activated expert modules.}
    \vspace{-1em}
    \label{fig:weight_sharing}
\end{figure*}


AutoMoE leverages the idea of Supernet training from prior works~\citep{onceforall,autodistill,hat} in Neural Architecture Search that were developed for standard non-MoE architectures. We extend Supernet training to the search space for MoE's by incorporating experts, gating and routing protocols. Typically, a Supernet consists of thousands of subnetworks that are all jointly trained via weight-sharing. %
The Supernet for {\sysname} is the largest sparsely activated MoE in the search space. It consists of the maximum number of experts ($M$) placed in every layer of the Transformer in both encoder and decoder. Each expert FFN has the maximum intermediate hidden size in the search space. Similar principles apply to the non-expert dense modules initialized with corresponding full dimension. 

The Supernet is trained with the following steps: (i) sample a candidate architecture randomly from the search space~\citep{spos}; (ii) train the sampled architecture by extracting the common portion of weights from different layers in the Supernet (i.e., by weight sharing) for one training step on the task; (iii) repeat steps (i) and (ii) until the training budget is exhausted. Once the Supernet training converges, we can obtain a quick accuracy estimate for a candidate architecture (i.e. subnetwork) by extracting its shared weights from the Supernet and evaluating on the validation set. 

The key challenge here is to build weight sharing techniques for MoE components, which include: (i) \textit{router}: a neural network that is trained to route each token (of `embedding size') in an incoming example to exactly one expert (out of $M$ experts) for top-$1$ routing; (ii) \textit{FFN expert}: a standard Transformer FFN block that has unique weights and is learned independently. {\sysname}'s expert layers follow the Switch Transformer~\citep{switchtransformers} specification. For subnetwork extraction from the Supernet, {\sysname} extracts front rows and front columns of the Supernet's router weight matrix, corresponding to the subnet design. For example, consider the Supernet's router to be designed for $4$ experts and $640$ embedding size with the shape of the router weight matrix as $4 \times 640$. Consider a sampled subnet during Supernet training to consist of $3 < 4$ experts and $512 < 640$ embedding size with the subnet's router matrix as $3 \times 512$. To populate this matrix, we extract the first $3$ rows and first $512$ columns from the Supernet's weight matrix (as illustrated in Figure~\ref{fig:weight_sharing} (a)). Such a weight sharing technique allows us to design hetegogeneous MoE architectures with varying number of experts in each Transformer layer. 

{\sysname} also extracts front rows and front columns from the weight matrices of each FFN expert from the Supernet, corresponding to the subnet design. For the previous example, assume the intermediate FFN size of each expert in the Supernet to be $3072$ (shape of weight matrix for first FFN layer is $3072 \times 640$ and second FFN layer is $640 \times 3072$). Assume the sampled subnet to be designed for $2$ experts with intermediate FFN size of one expert to be $2048$ while the other to be $1024$. For the first expert, the weight matrices of the subnet of shape $2048 \times 512$ (Input) and $512 \times 2048$ (Output) are extracted from the first $2048$ rows, $512$ columns (Input) and first $512$ rows, $2048$ columns (Output) of the corresponding Supernet weights. For the second expert, the weight matrices of shape $1024 \times 512$ (Input) and $512 \times 1024$ (Output) are extracted from the first $1024$ rows, $512$ columns (Input) and first $512$ rows, $1024$ columns (Output) of the corresponding Supernet weights. This example is illustrated in Figure~\ref{fig:weight_sharing} (b). The subnet extraction technique does not extract weights from the third and fourth experts of the Supernet as the subnet is designed to have only two experts (not shown in the figure). Such a weight sharing technique allows us to design architectures with varying intermediate FFN size for each expert. Additional techniques for improving expert capacity such as stacking FFNs, and techniques for improving Supernet performance with sandwich sampling~\citep{sandwich}, inplace knowledge distillation~\citep{sandwich}, gradient conflict reduction~\citep{nasvit} are left for future work.     

\subsection{Searching for Efficient MoE Subnetwork with Computational Constraint}
\label{sec:evo_search}
{\sysname} search is based on an evolutionary algorithm that takes the hardware computational constraint (e.g., CPU latency $\le 600$ms) as input and aims to identify the MoE subnetwork from the Supernet which achieves maximum accuracy for the task while satisfying the constraint. The algorithm works by sampling an initial set of MoE candidate architectures randomly from the Supernet; evolving the top architectures iteratively by mutation; followed by crossover; until the search iterations are exhausted. Candidate MoE architectures are easily ranked by the Supernet  performance estimator based on the validation score for the task. Latency estimate for each architecture is obtained by measuring the latency directly on the target device. The standard approach measures gold latency for forward propagation of a batch of examples for a large number (e.g., 300) of passes and then computes the truncated mean (after removing bottom and top 10\% outlier latencies). This latency estimation can be costly given the large space of candidate architectures. To overcome this challenge, {\sysname} uses \textit{partially gold latency}, which is obtained by forward propagation of a batch of examples for a small number (e.g., $100$) of passes and then computing truncated mean. After the search is completed, the MoE architecture with the highest performance is selected as the optimal one.

\subsection{Training Efficient MoE Sub-Transformer}
\label{sec:train_eff_sntw}
Once the optimal MoE architecture is identified, 
we train the model weights for the final architecture to convergence for the same number of training steps as our baseline models  
for a fair comparison.


%% file: experiments.tex
\section{Experiments}
\label{sec:expertiments}

\noindent{\bf Datasets and evaluation metrics.}
\label{sec:data_eval}

\begin{table}
\scriptsize
\begin{center}
\begin{tabular}{p{0.2in}p{0.2in}p{0.2in}p{0.2in}p{0.2in}p{0.2in}p{0.2in}} \toprule
\textbf{Dataset} & \textbf{Year} & \textbf{Source Lang} & \textbf{Target Lang} & \textbf{\#Train} & \textbf{\#Valid} & \textbf{\#Test} \\ \midrule
WMT & 2014 & English (en) & German (de) & 4.5M & 3000 & 3000 \\
WMT & 2019 & English (en) & German (de) & 43M & 2900 & 2900 \\
WMT & 2014 & English (en) & French (fr) & 35M & 26000 & 26000 \\
\bottomrule
\end{tabular}
\vspace{-1em}
\caption{Machine translation benchmark data.}
\vspace{-2em}
\label{tab:dataset_statistics}
\end{center}
\end{table}

\begin{table*}[t]
\scriptsize
\begin{center}
\begin{tabular}{p{1in}p{0.8in}p{0.6in}p{0.6in}p{0.5in}p{0.3in}p{0.5in}p{0.6in}} \toprule 
\textbf{Dataset} & \textbf{Network} & \textbf{\#Active Params (M)} & {\bf Sparsity (\%)} & \textbf{FLOPs (G)} & \textbf{BLEU} & \textbf{GPU hours} & {\textbf{Latency (ms)}} \\ \midrule

\multicolumn{6}{l}{\textbf{WMT’14 En-De}} \\ 
Transformer-Big & Dense & 176 & 0 & 10.6 ($1\times$) & {28.4} & 184 & {\multirow{2}{*}{2199 ($1\times$)}} \\
{SwitchTransformer-Big} & {Sparse} & {176} & {36} & {10.6} ($1\times$) & {{\bf 28.8}} & {236} &  \\
Evolved Transformer & NAS over Dense & 47 & 0 & {\bf 2.9} ($3.7\times$) & 28.2 & 2,192,000 & {-}  \\
HAT & NAS over Dense & 56 & 0 & 3.5 ($3\times$) & 28.2 & 264 & {669} ($3.3\times$) \\
{Random Search} & {NAS over Sparse} & {42} & {21} & {2.2} ($4.8\times$) & {27.3} & {126} & {416} ($5.3\times$) \\
\textcolor{blue}{AutoMoE (6 Experts)} & \textcolor{blue}{NAS over Sparse} & \textcolor{blue}{{\bf 45}} & \textcolor{blue}{62} & \textcolor{blue}{{\bf 2.9} ($3.7\times$)} & \textcolor{blue}{28.2} & \textcolor{blue}{224} & \textcolor{blue}{504 ($4.4\times$)} \\  \midrule

\multicolumn{6}{l}{\textbf{WMT’14 En-Fr}} \\ 
Transformer-Big & Dense & 176 & 0 & 10.6 ($1\times$) & 41.2 & 240 & {\multirow{2}{*}{2199 ($1\times$)}} \\
{SwitchTransformer-Big} & {Sparse } & {176} & {36} & {10.6} ($1\times$) & {{\bf 42.3}} & {234} &  \\
Evolved Transformer & NAS over Dense & 175 & 0 & 10.8 ($1\times$) & 41.3 & 2,192,000 & - \\
HAT & NAS over Dense & 57 & 0 & 3.6 ($2.9\times$) & 41.5 & 248 & {723} ($3\times$) \\
{Random Search} & {NAS over Sparse} & {42} & {21} & {2.2} ($4.8\times$) & {40.3} & {130} & {416} ($5.3\times$) \\
\textcolor{blue}{AutoMoE (6 Experts)} & \textcolor{blue}{NAS over Sparse}  & \textcolor{blue}{{\bf 46}} & \textcolor{blue}{72} & \textcolor{blue}{{\bf 2.9} ($3.7\times$)} & \textcolor{blue}{{41.6}} & \textcolor{blue}{\multirow{2}{*}{236}} & \textcolor{blue}{547 ($4\times$)  } \\ 
AutoMoE (16 Experts) & NAS over Sparse & 135 & 65 & 3.0 ($3.5\times$) & { 41.9} & & {672} ($3.3\times$) \\  
\midrule

\multicolumn{6}{l}{\textbf{WMT’19 En-De}} \\
Transformer-Big & Dense  & 176 & 0 & 10.6 ($1\times$) & { 46.1} & 184 & {\multirow{2}{*}{2199 ($1\times$)}}  \\
{SwitchTransformer-Big} & {Sparse } & {176} & {36} & {10.6} ($1\times$) & {{\bf 47.0}} & {223} &  \\
HAT & NAS over Dense  & 63 & 0 & 4.1 ($2.6\times$) & 45.8 & 264 & {758} ($2.9\times$) \\
{Random Search} & {NAS over Sparse} & {42} & {21} & {2.2} ($4.8\times$) & {43.7} & {126} & {416} ($5.3\times$) \\
\textcolor{blue}{AutoMoE (2 Experts)} & \textcolor{blue}{NAS over Sparse}  & \textcolor{blue}{{\bf 45}} & \textcolor{blue}{41} & \textcolor{blue}{{\bf 2.8} ($3.8\times$)} & \textcolor{blue}{45.5} & \textcolor{blue}{\multirow{2}{*}{248}} & \textcolor{blue}{558 ($3.9\times$)}  \\ 
AutoMoE (16 Experts) & NAS over Sparse  & 69 & 81 & 3.2 ($3.3\times$) & {45.9} &  & {656 ($3.3\times$)} \\

\bottomrule
\end{tabular}
\vspace{-.3em}
\caption{Comparison of AutoMoE vs. baselines with Pareto-optimal architectures highlighted in \textcolor{blue}{blue color}. We report active model parameters, and sparsity measured as non-active parameters as a percentage of total parameters. We train all baselines and {\sysname} for the same $40K$ training steps for fair comparison to report BLEU\footnotemark. Training time (with search, if applicable) is reported in hours for one Nvidia V$100$ GPU. Inference latency is measured in Intel Xeon CPU. {\sysname} significantly reduces FLOPs and latency with parity in BLEU, on aggregate, over NAS methods in dense search space (e.g., $1.3\times$ and $2.4\times$ FLOPs reduction and speedup over HAT and Evolved Transformer). {\sysname} with Random Search obtains the best speedup but results in significant regression in BLEU.}
\label{tab:nasmoe_vs_baseline}
\vspace{-1.7em}
\end{center}
\end{table*}

We evaluate {\sysname} on standard machine translation benchmarks: WMT'14 En-De, WMT'14 En-Fr and WMT'19 En-De 
with dataset statistics in Table~\ref{tab:dataset_statistics}. We use pre-processed datasets and evaluation setup from ~\cite{hat}. We report BLEU score~\citep{bleuscore} as a performance metric with beam of size $5$ and a length penalty of $0.6$ (for WMT).

\noindent{\bf Baselines.}
We compare {\sysname} against both manually designed and NAS-searched architectures. 

For {\bf manual baselines}, we consider: (a) densely activated Transformers~\cite{NIPS2017_3f5ee243} with no experts; (b) sparsely activated MoE with homogeneous experts (i.e. same number and FFN size) placed in every other layer~\citep{switchtransformers,kim_etal_21,stochasticexperts,glam,artetxe_arxiv21}. 

For {\bf NAS baselines}, we consider (c) HAT~\citep{hat}, which is a Supernet-based state-of-the-art NAS framework for identifying efficient dense sub-Transformers for neural machine translation (same task setting as ours); and (d) Evolved Transformer~\citep{evotrans} which is one of the earlier works on finding efficient dense sub-Transformers with evolution-based architecture search. {\em Note that both the NAS baselines apply only to dense non-MoE transformers, and {\sysname} is the first work to leverage NAS to identify efficient sparse MoE sub-transformers.} Finally, we consider (e) {\sysname} with Random Search (typically treated as a strong baseline for NAS) that samples an MoE subnetwork (given latency constraints) randomly from {\sysname} search space and trains it till convergence.


\noindent{\bf Training configurations and search space.} All the baselines and {\sysname} including the Supernet and final model are trained with the same setting for fair comparison. All the models are trained for $40K$ steps, with a warmup of $10K$ steps from $10^{-7}$ to $10^{-3}$ and use cosine annealing to $10^{-7}$ for the rest of the steps. 
All models are trained using fairseq toolkit~\citep{fairseq} with an effective batch size of $524K$ tokens on $16$ V$100$ GPUs. All the NAS baselines have the same search space for dense Transformer modules (e.g., number of decoder layers, q-k-v dimension, attention heads, etc.) with {\sysname} further incorporating MoE relevant aspects (e.g., experts, gating, routing, etc.) in the search space. The number of encoder layers is kept fixed for all the NAS baselines including {\sysname} since the latency is primarily determined by the decoders for autoregressive generation (as we discuss in Section~\ref{subsec:analysis}).

\begin{figure*}[t!]
    \centering
    \begin{subfigure}[t]{0.4\textwidth}
        \centering
        \includegraphics[height=1.2in, width=1.4in]{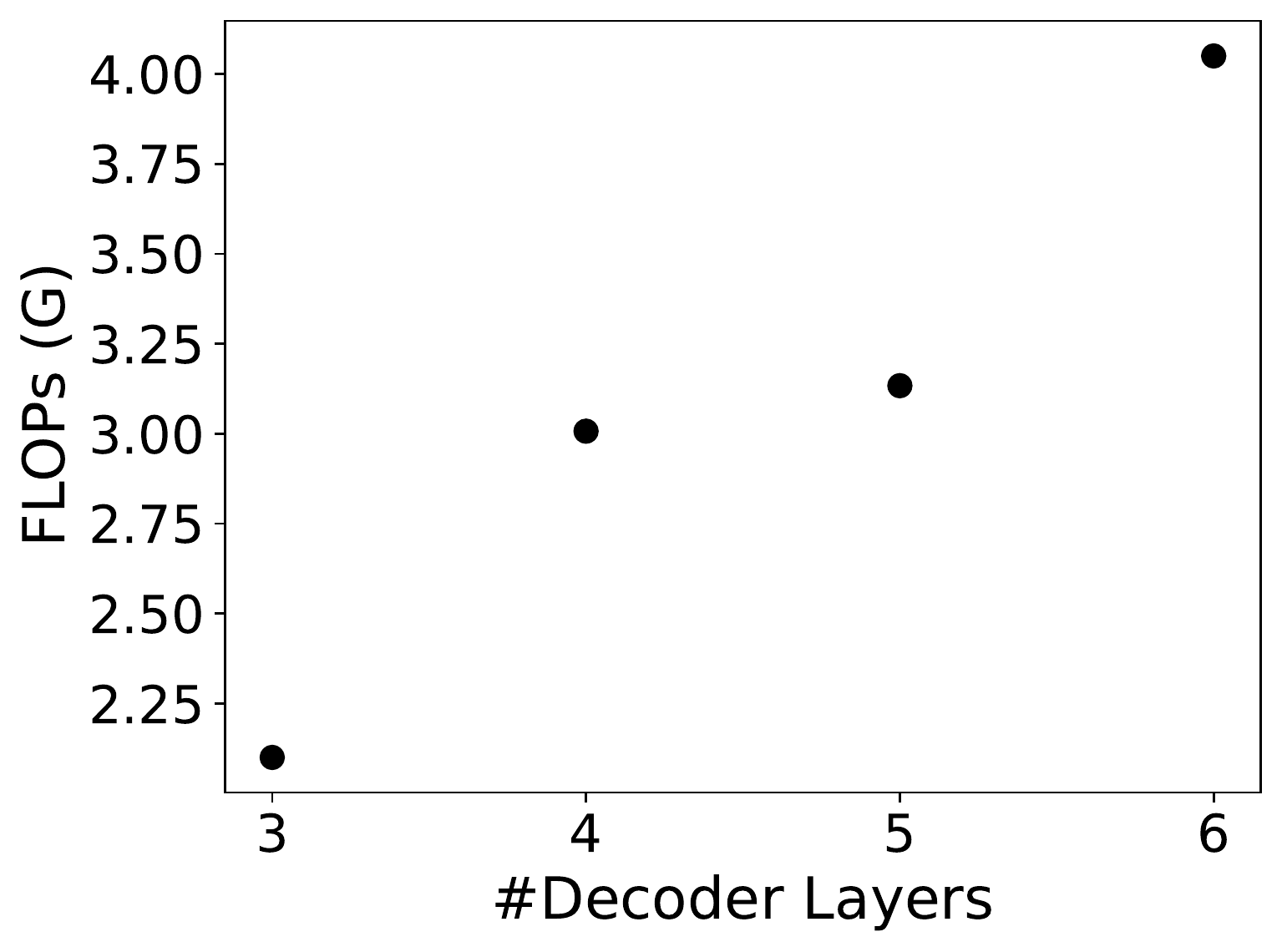}
        \caption{\#Decoder layers vs. FLOPs.}
    \end{subfigure}%
    ~ 
    \begin{subfigure}[t]{0.4\textwidth}
        \centering
        \includegraphics[height=1.2in, width=1.4in]{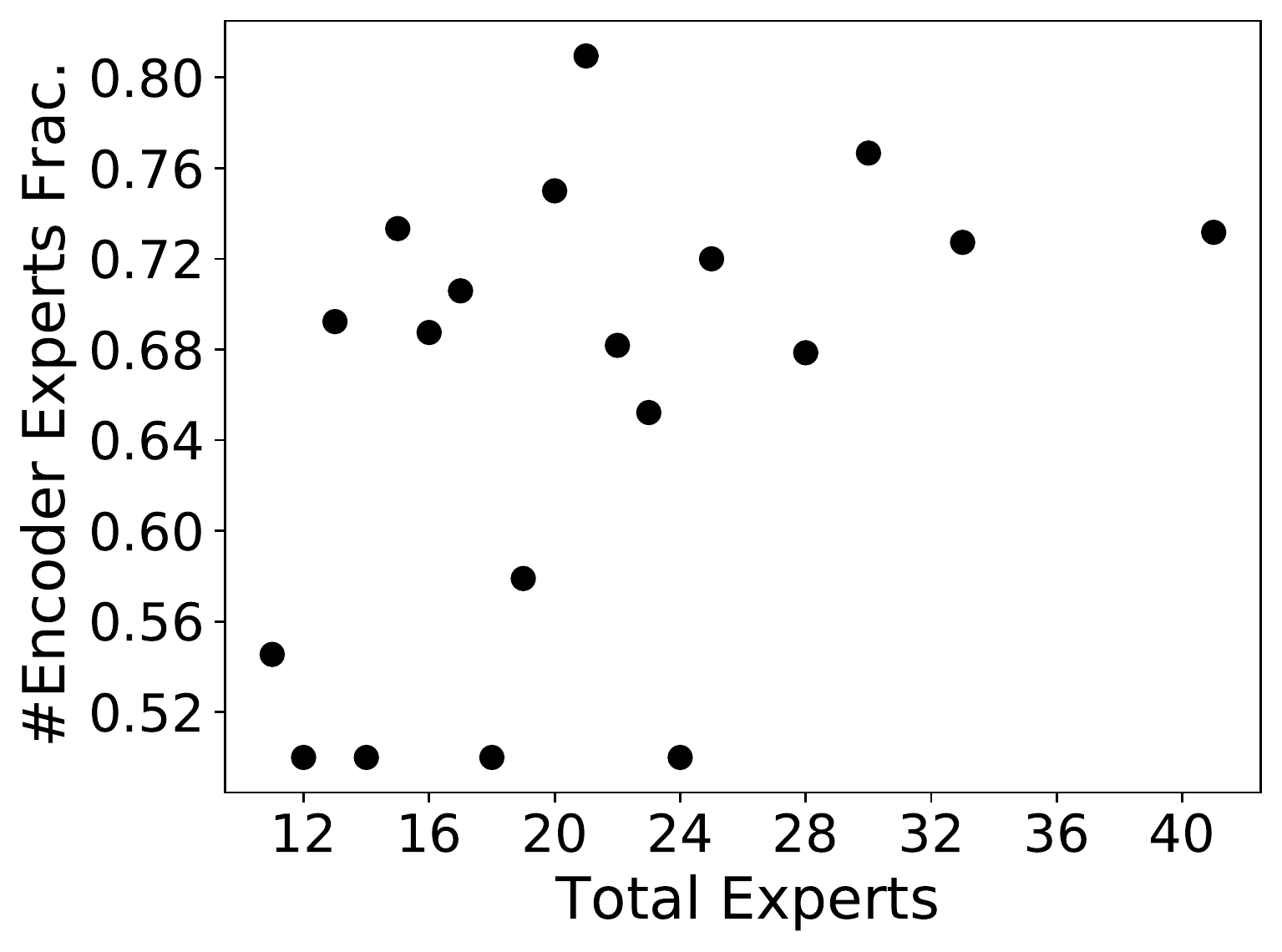}
        \caption{Encoder experts as ratio of total experts.}
    \end{subfigure}
    
    \begin{subfigure}[t]{0.33\textwidth}
        \centering
        \includegraphics[height=1.2in, width=1.4in]{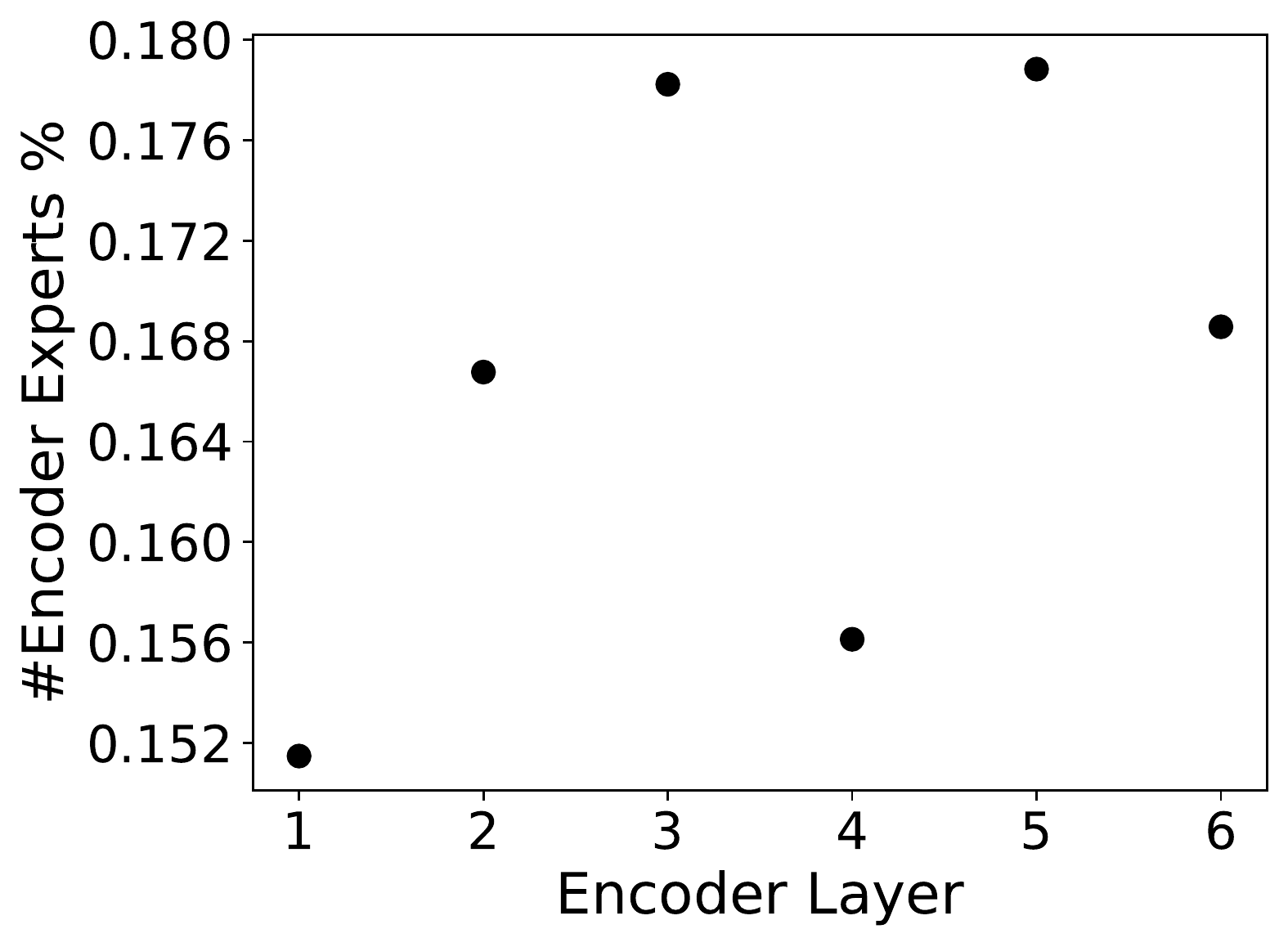}
        \caption{Encoder layer vs. \#Expert (\%).}
    \end{subfigure}%
    ~
    \begin{subfigure}[t]{0.33\textwidth}
        \centering
        \includegraphics[height=1.2in, width=1.4in]{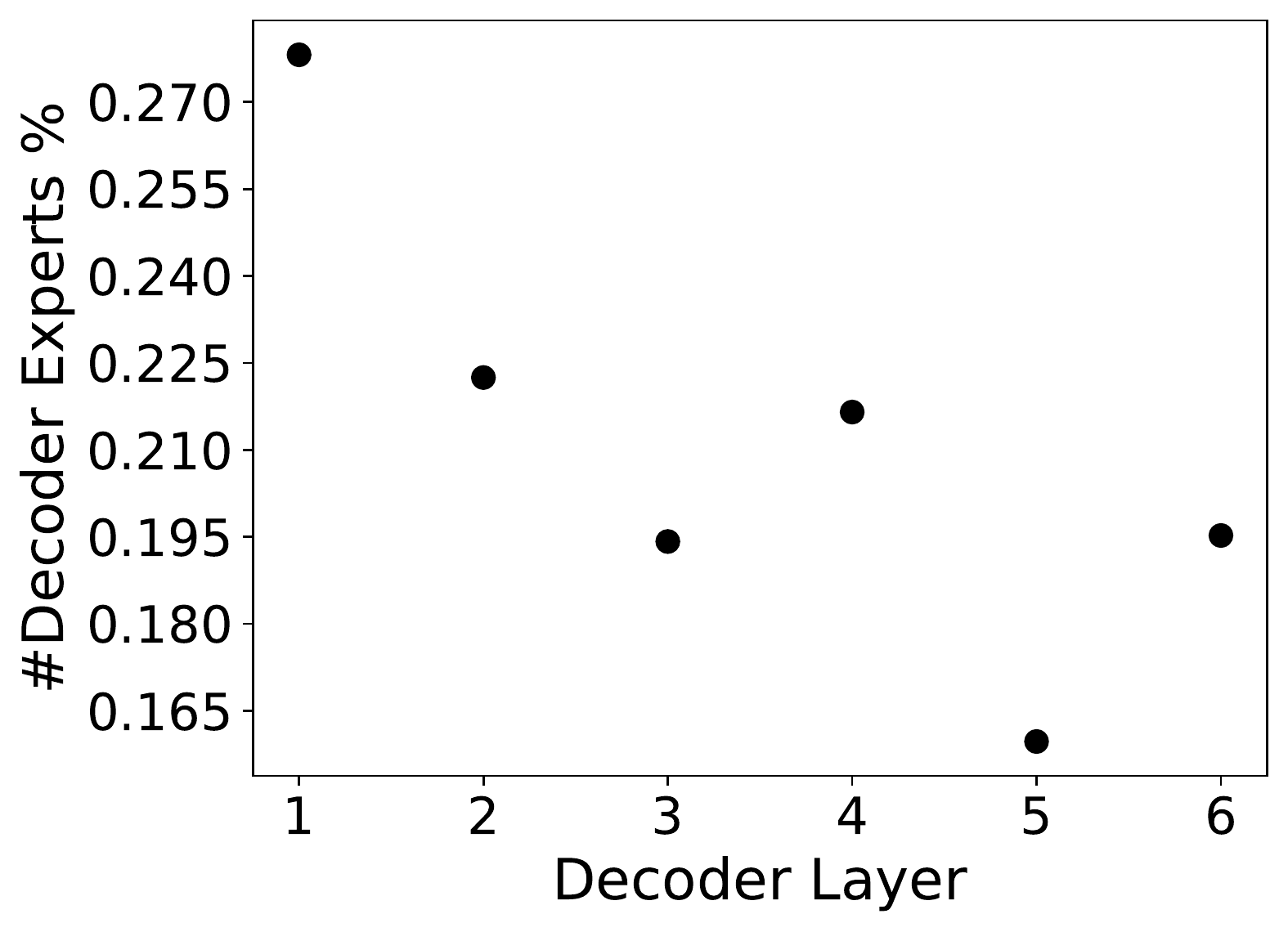}
        \caption{Decoder layer vs. \#Expert (\%).}
    \end{subfigure}%
    ~
    \begin{subfigure}[t]{0.33\textwidth}
        \centering
        \includegraphics[height=1.2in, width=1.4in]{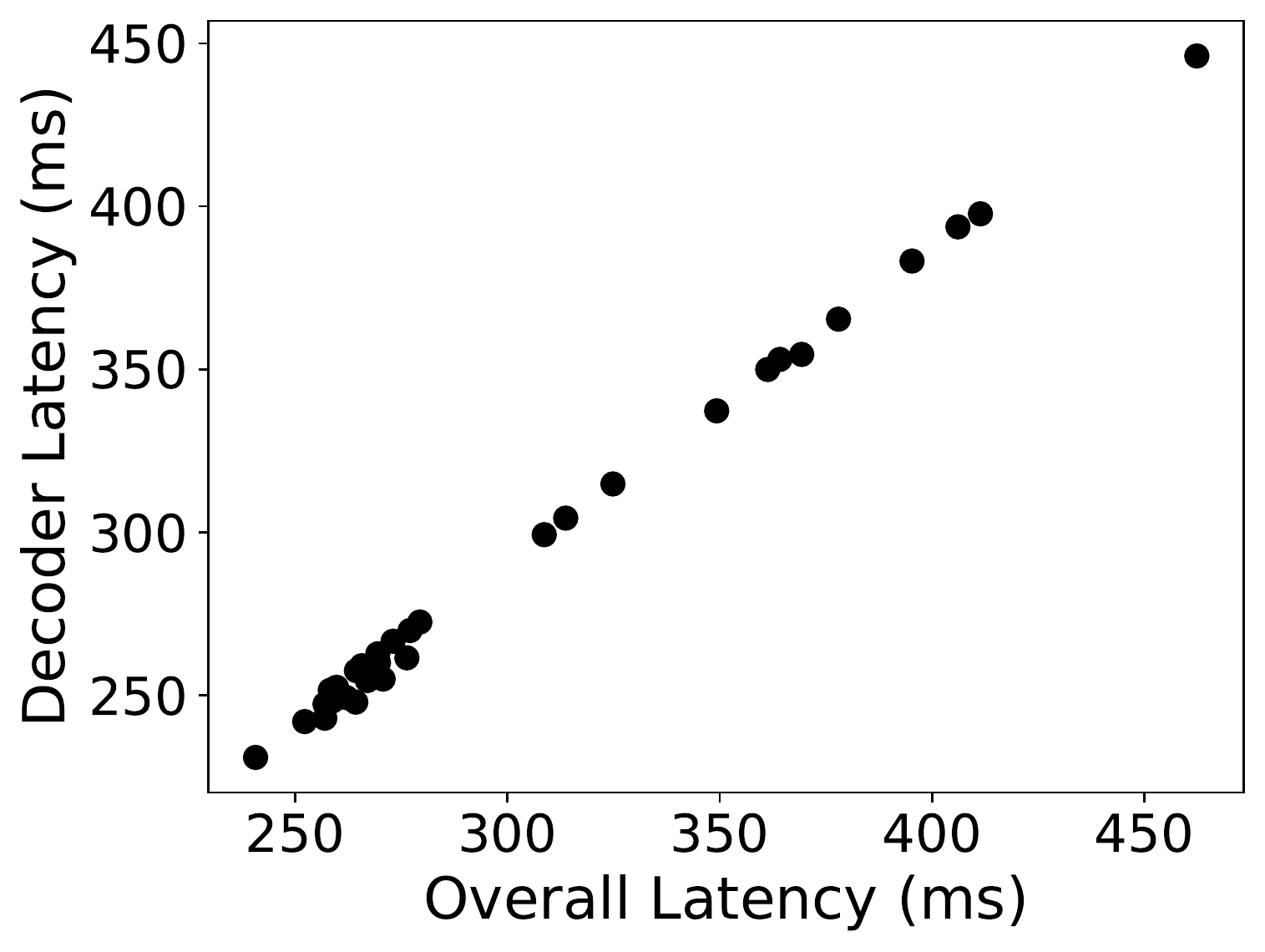}
        \caption{Overall vs. Decoder latency.}
    \end{subfigure}%
    \vspace{-0.5em}
    \caption{Architecture analysis for {\sysname}-generated MoEs. We sample several architectures from the Pareto for {\sysname} subnets, and report aggregate statistics in terms of the impact on different computational metrics.}
    \vspace{-1em}
\label{fig:architecture_analysis}
\end{figure*}

\noindent{\bf Evolutionary search setup.}
For performance estimation, we monitor the validation loss of subnets on the NMT task.
We compute latency by measuring the time taken to perform translation from a source sentence to a target sentence with same desired input/output length ($30$ for WMT) and original beam settings (see Section~\ref{sec:data_eval}) on target device (Intel Xeon CPU). We measure latency $300$ times for gold (to report final metrics) and $100$ times for partially gold (during evolutionary search) respectively; discard top and bottom 10\% (outlier latency) and compute mean of the rest. Hyper-parameter settings for evolutionary search include: $15$ as iterations, $125$ as population size, $25$ as parents' size, $50$ as mutation population size with mutation probability of $0.3$ and $50$ as crossover population size. Unless otherwise stated, latency constraint for all experiments is set to $600ms$.

%% file: results.tex
\section{Results}
\label{sec:results}

\begin{table*}[htb]
\scriptsize
\begin{center}
\begin{tabular}{p{0.8in}p{0.8in}p{1.8in}p{0.8in}p{1.2in}} \toprule
\textbf{Model} & \multicolumn{2}{c}{\textbf{Encoder}} & \multicolumn{2}{c}{\textbf{Decoder}} \\ 
\textbf{Dataset} & \textbf{\#Experts per layer} & \textbf{Expert FFN Inter Size} &  \textbf{\#Experts per layer} & \textbf{Expert FFN Inter Size} \\ \midrule
\textbf{Std-expert} \\
WMT'14 En-De & 5-1-1-1-2-1 & 3072-3072-3072-3072-2048-3072 & 1-1-1-1 & 3072-3072-3072-3072 \\   
WMT'14 En-Fr & 1-4-2-6-5-5 & 3072-3072-3072-3072-3072-3072 & 2-1-1-3 & 3072-3072-3072-3072 \\
WMT'19 En-De & 1-1-2-1-2-1 & 3072-3072-3072-3072-3072-2048 & 1-1-1-2 & 3072-3072-3072-3072 \\ \midrule
\textbf{Fract-expert} \\
WMT'14 En-De & 3-2-3-4-1-3 & [2048-3072-2048]-[3072-1024]-[3072-3072-1024]-[3072-1024-3072-2048]-3072-[3072-1024-3072] & 3-1-1-1 & [3072-1024-2048]-3072-3072-3072 \\   
WMT'14 En-Fr & 6-2-3-4-4-5 & [2048-1024-2048-1024-1024-3072]-[2048-2048]-[3072-3072-2048]-[3072-3072-2048-3072]-[3072-1024-1024-2048]-[2048-3072-3072-2048-2048] & 2-1-4-2 & [3072-3072]-3072-[3072-3072-3072-2048]-[3072-2048] \\
WMT'19 En-De & 2-3-1-2-6-1 & [3072-3072]-[3072-3072-3072]-3072-[3072-2048]-[3072-1024-2048-3072-1024-2048]-3072 & 2-4-1-1 & [3072-3072]-[3072-1024-2048-3072]-3072-3072 \\ \bottomrule
\end{tabular}
\caption{{\sysname}-generated Pareto-optimal architectures for different datasets. FFN intermediate sizes for fractional experts (i.e. varying expert sizes within each layer) are enclosed within square brackets.}
\label{tab:nasmoe_designs}
\end{center}
\end{table*}



\subsection{AutoMoE vs. Baseline Performance}
Table~\ref{tab:nasmoe_vs_baseline} presents a comparison of 
{\sysname} with baselines on several computational metrics and task performance. We report the number of parameters without embedding weights, and FLOPs without the last decoding layer for all the models, consistent with~\cite{hat} evaluation. \sysname-generated sparse MoE sub-Transformers obtain $4\times$ reduction in FLOPs over both manually designed (densely activated) Transformer-Big, and (sparsely activated) MoE SwitchTransformer-Big with experts in every layer, and equivalent inference speedups on CPU. Compared to NAS baselines like Evolved Transformer~\cite{evotrans} and HAT~\cite{hat} that generate densely activated sub-Transformers, {\sysname} improves on FLOPs and latency by $2.4\times$ and $1.3\times$ respectively with parity in BLEU score on aggregate. Notably, Supernet-based {\sysname} and HAT have massively reduced amortized training cost (GPU hours) compared to Evolved Transformer with progressive evolutionary search. {\sysname} with Random Search, a strong NAS baseline, obtains the best speedup but with significant performance regression.

Compared to all other models (both dense and sparse), we observe {\sysname} to generate networks with high sparsity resulting in massively reduced active parameters and FLOPs. For the NAS models, we train the top-$2$ sub-Transformers in the Pareto and report the one with the best trade-off in BLEU vs. FLOPs on the validation set. Maximum experts for the best performance vary for different tasks, with $6$ experts for WMT'14 En-De, $16$ experts for WMT'14 En-Fr and WMT'19 En-De -- given the latter two datasets are $10\times$ larger than the former.

\subsection{Analysis}
\label{subsec:analysis}
\noindent{\bf Decoder layers vs. FLOPs.}
Figure~\ref{fig:architecture_analysis} (a) shows the average FLOPs for several {\sysname} architectures with different decoder layers as obtained during our search (varying from $3$ to $6$) from the Pareto, and baseline models. Notice that the FLOPs increase with increase in decoder layers, given the auto-regressive nature of NMT tasks which require generating tokens sequentially. In contrast to manually designed Transformers with $6$ decoder layers (both dense and sparsely activated MoE variants), {\sysname}- and HAT-searched architectures reduce the number of decoder layers with a resulting decrease in both FLOPs and latency. This is also evident in Figure~\ref{fig:architecture_analysis} (e) which shows that decoder latency dominates the total inference latency for all the models by more than $90\%$. 

\footnotetext{We use same hyper-parameters for all models with no tuning (provided in code). Given $40K$ training steps for each model and no tuning, MoE numbers may not be comparable to SOTA numbers which typically train for more steps. HAT and Evol. Transformer numbers are reported from~\cite{hat}. We follow their evaluation and reporting protocol.}

\noindent{\bf Expert distribution in encoder vs. decoder.}
Figure~\ref{fig:architecture_analysis} (b) plots the number of encoder experts as ratio of total experts for {\sysname}-generated sub-Transformers. We observe that {\sysname}  assigns significantly larger number of experts to encoder as compared to the decoder. As a result, encoders have much higher capacity (i.e., encoder parameters as a proportion of overall parameters) than decoders. This correlates with the earlier observation that models with higher encoder layers compared to decoder layers enjoy better latency-performance trade-off~\citep{kasai2021deep}. Our findings from {\sysname} designed architectures indicate that the number of layers and experts are two knobs that jointly help in modulating encoder capacity and decoder latency to design efficient MoE. 

\noindent{\bf Expert distribution in different layers.}
Figures~\ref{fig:architecture_analysis} (c) and (d) show the percentage of experts allocated to different layers for encoders and decoders -- averaged over several sampled architectures from {\sysname} Supernet. Notice that the middle encoder layers ($3^{rd}, 5^{th}$) are  allocated the maximum number of experts, while the first layer receives the least. The trend reverses for decoder, with the first layer receiving most experts with gradual reduction in expert allocation. This is also consistent with keeping decoders light by dropping layers to reduce latency; while compensating for the reduced capacity with increased experts in the first few layers.

\begin{table}
\scriptsize
\begin{center}
\begin{tabular}{lccc} \toprule
\textbf{Search Constraint} & \textbf{BLEU} & \textbf{FLOPs (G)} & \textbf{Latency (ms)}  \\ \midrule
\multicolumn{4}{l}{\textbf{Latency  $\mathbf{\leq}$ 200ms}}  \\
{HAT} & {41.45} & {3.6} & {212} \\
{\sysname} (2 Experts) & 41.23 & 2.9 & 176  \\
{\sysname} (4 Experts) & 41.22 & 3.0 & 198 \\ 
\midrule
\multicolumn{4}{l}{\textbf{FLOPs  $\mathbf{\leq}$ 3 GFLOPs}}  \\
{HAT}  & {40.89} & {3.0} & {158} \\
{\sysname} (2 Experts)  & 41.09 & 3.0 & 216 \\
{\sysname} (4 Experts) & 41.10 & 3.0 & 229 \\

\bottomrule
\end{tabular}
\caption{Impact of latency and FLOPs constraints on WMT'14 En-Fr dataset. Latency is computed on $1$ NVIDIA V100 GPU.}
\label{tab:nasmoe_diff_constraint_types}
\end{center}
\end{table}

\noindent{\bf Latency vs. FLOPs as constraint for search.}
Table~\ref{tab:nasmoe_diff_constraint_types} presents the impact of latency and FLOPs as computational constraints on the performance-efficiency trade-off. Constraining FLOPs results in models that fully exhaust the FLOPs budget; while leading to higher latency. On the other hand, constraining latency tends to under-utilize the budget leading to relatively superior FLOPs and latency,   
providing a stricter control.

\noindent{\bf Pareto-optimal {\sysname} generated MoE architectures.} Table~\ref{tab:nasmoe_designs} shows sparsely activated MoE architectures designed by two variants of {\sysname} (`std-expert': expert FFN size same in each layer and variable across; `fract-expert': fully heterogeneous expert size) for different datasets with the best trade-off in BLEU vs. latency. On aggregate $71\%$ of the experts are allocated to the encoder compared to the decoder. Meanwhile, $70\%$ of the expert layers in `fract-expert' architectures have $2$ or more experts, out of which more than $75\%$ of the expert layers have varying capacities (i.e., experts with different FFN intermediate size). Figures~\ref{fig:arch_designs_wmt14ende},~\ref{fig:arch_designs_wmt14enfr},~\ref{fig:arch_designs_wmt19ende} in Appendix show full architecture (embedding size, layers, heads, experts, placement, sizes, etc.) of {\sysname} subnets on WMT14 En-De, WMT14 En-Fr and WMT19 EN-De respectively.

\begin{table}
\scriptsize
\begin{center}
\begin{tabular}{p{2.0in}p{0.2in}p{0.3in}} \toprule
\textbf{Search Space Variation} & \textbf{BLEU} & \textbf{FLOPs}  
\\ \midrule
HAT  & 28.2 & 3.5G \\
{\bf AutoMoE (2 Experts) w/ fixed encoder layers} & {\bf 28.2} & {\bf 2.9G}\\
\midrule
\multicolumn{3}{p{2in}}{\textbf{Varying number of encoder layers}}\\
HAT w/ \#Encoder-Layers $\in$ \{1-6\} & 28.1 & 3.4G \\
AutoMoE (2 Experts) w/ \#Encoder-Layers $\in$ \{1-6\} & 28.3 & 3.7G \\
\midrule
\multicolumn{3}{p{2in}}{\textbf{AutoMoE (2 Experts) w/ manually designed homogeneous experts}} \\ 
1-2-1-2-1-2 & 28.3 & 3.5G \\
1-1-1-2-2-2 & 28.3 & 3.8G \\
2-2-2-1-1-1 & 28.3 & 3.1G \\
\midrule

{\bf \sysname\ w/ Identity Expert} FFN size $\in$ \{0, 3072\}  & 28.1 & 2.7G \\
  \bottomrule
\end{tabular}
\vspace{-1em}
\caption{Variations in \sysname's search space on WMT'14 En-De dataset.}
\label{tab:nasmoe_design_variations}
\end{center}
\vspace{-2.5em}
\end{table}

\noindent{\bf MoE Search space variations.}
Table~\ref{tab:nasmoe_design_variations} presents the impact of search space choices on MoE efficiency and performance trade-off. The first variation is to make `\#Encoder Layers' an elastic search dimension. Note that both HAT and {\sysname} consider the number of encoder layers to be fixed (refer to Table~\ref{tab:design_space}). We observe that varying encoder layers has a relatively poor trade-off on model performance vs efficiency as compared to varying decoder layers, re-inforcing our prior observations on the importance of encoder capacity and depth.


In the second variation (see third major row), we fix the expert architecture (with $2$ experts manually placed uniformly) in the search space and only search for standard Transformer hyper-parameters. Observe that  {\sysname}-designed models have better FLOPs than such manually designed ones. 


The last variation introduces identity or dummy experts (i.e.,  expert with $0$ intermediate FFN size, equivalent to identity operation). This explores the idea that we can {\em skip} the computation for some of the tokens based on context rather than always forcing them through an FFN. We observe identity experts to marginally hurt the performance but significantly reduce FLOPs (see last major row). 

%% file: conclusion.tex
\section{Conclusion}
\label{sec:conclusion}

{\sysname} is the first framework to design heterogeneous MoE's under computational constraints. It supports adaptive computation i.e. variable compute for different inputs with variable-size experts. It leverages NAS to explore a heterogeneous search space with variable number of experts, sizes, and placement choices; alongside other standard Transformer architectural hyper-parameters. {\sysname} generated MoE subnetworks reduce FLOPs and latency over both manually designed and NAS-searched architectures on benchmark MT tasks.



%% file: limitations.tex
\section{Limitations}
\label{sec:limitations}

Given our focus on finding efficient MoE models {\em under computational constraints}, {\sysname} search space and evaluation has been restricted in scale to big-sized Transformer models for benchmark MT tasks. A natural extension of this work is to explore the limits of MoE models like SwitchTransformers~\citep{switchtransformers} and GShard~\citep{gshard} that are significantly larger containing billions to trillions of parameters; as well as designing sparse and transferable efficient expert models~\citep{stmoe} for diverse types of tasks like reasoning, summarization and understanding.

The limitations of this work are as follows:
\begin{enumerate}
 \item Sandwich sampling~\citep{sandwich}, inplace knowledge distillation~\citep{sandwich}, and gradient conflict reduction~\citep{nasvit} are popular techniques to improve the training procedure of supernet. It would be interesting to study the impact of these techniques to improve {\sysname}'s supernet.
 \item {\sysname} uses the hidden dimension of intermediate feedforward network (FFN) to modulate the capacity of each expert. It would be interesting to study other techniques to modulate expert capacity such as stacking variable number of hidden layers in FFN.
 \item The backbone of {\sysname}'s supernet uses Switch Transformer, which adds FFN based expert layers and routes each token to exactly one expert (top-1 routing). It would be interesting to: (i) search for the number of tokens to route, and (ii) search for the Transformer component (e.g., FFN, self-attention projection layers, LayerNorm) to add expert layers.
 \item {\sysname}'s search space contains classical Transformer components such as multi-head attention and FFN layers. It would be interesting to add components that are efficient by design such as convolutional layer, FLASH~\cite{flashattention}, and g-MLP~\cite{gmlp}.
\end{enumerate}

%% file: ack.tex
\section*{Acknowledgements}\label{sec:acknow}
MAM acknowledges support from Canada Research Chairs (CRC), the Natural Sciences and Engineering Research Council of Canada (NSERC; RGPIN-2018-04267), Canadian Foundation for Innovation (CFI; 37771), and Digital Research Alliance of Canada.\footnote{\href{https://alliancecan.ca}{https://alliancecan.ca}} Lakshmanan's research was supported in part by a grant from NSERC (Canada).  

%% file: appendix.tex
\section{Appendix}
\label{sec:appendix}

\subsection{Full Architecture Design}
\begin{figure}
\centering
\includegraphics[width=3.0in,height=8.0in]{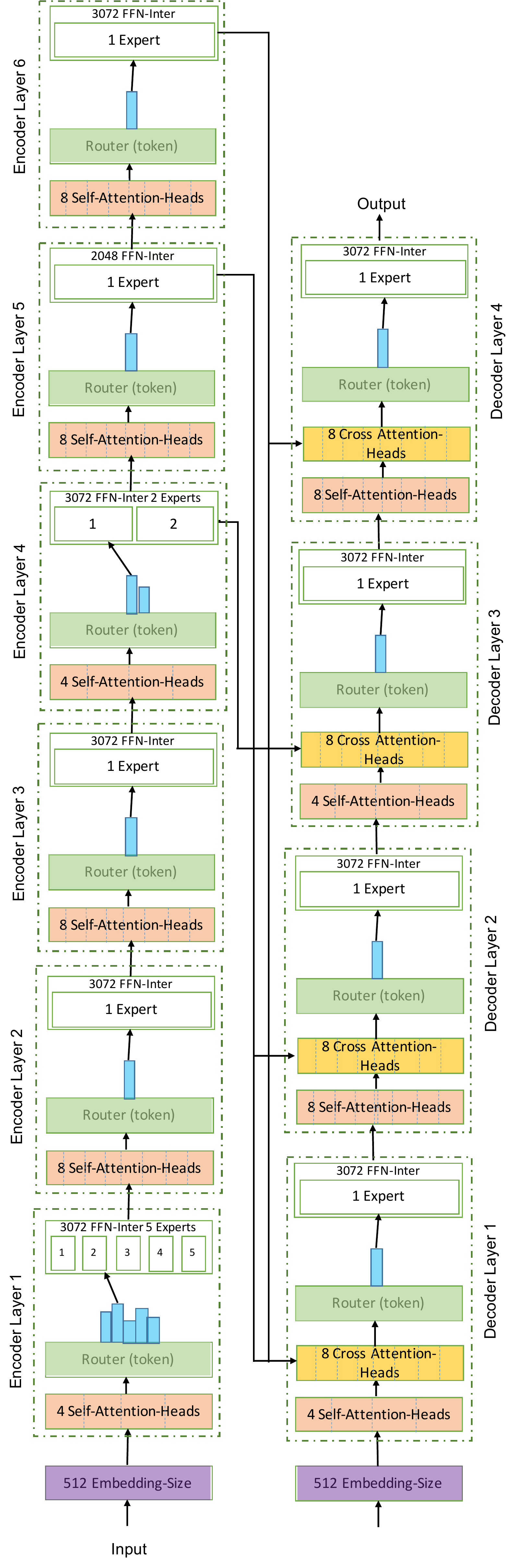}
\vspace{-0.8em}
\caption{{\sysname}-generated architecture for WMT'14 En-De.}
\label{fig:arch_designs_wmt14ende}
\vspace{-2em}
\end{figure}

\begin{figure}
\centering
\includegraphics[width=3.0in,height=8.0in]{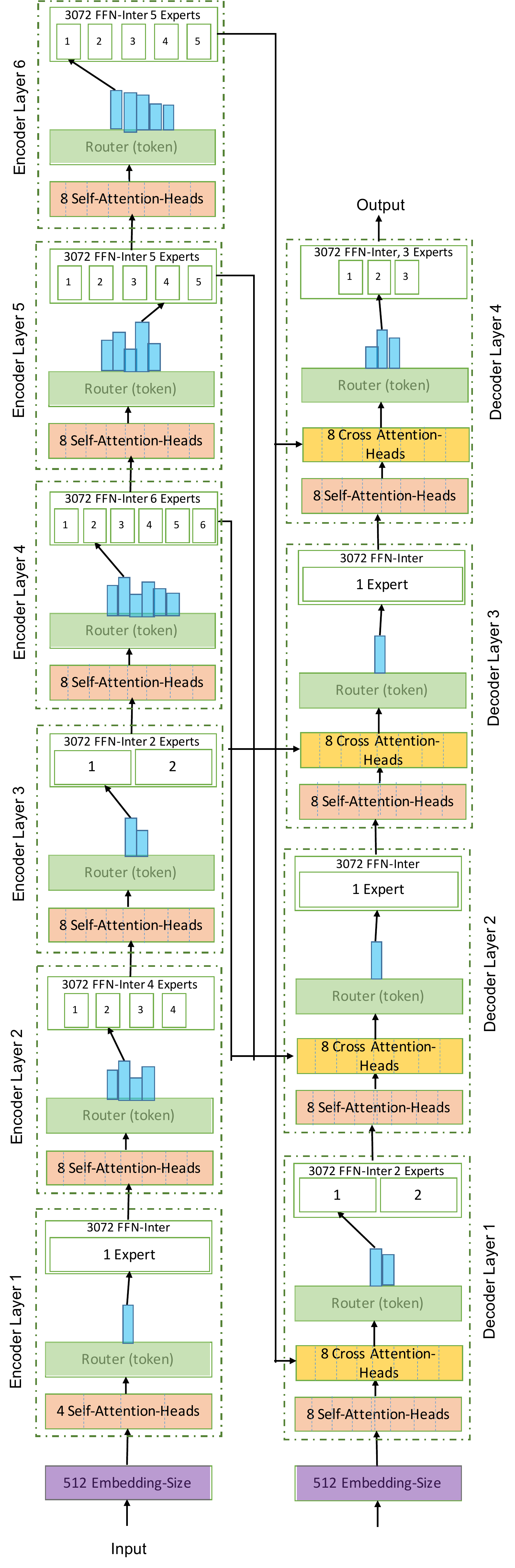}
\vspace{-0.8em}
\caption{{\sysname}-generated architecture for WMT'14 En-Fr.}
\label{fig:arch_designs_wmt14enfr}
\vspace{-2em}
\end{figure}

\begin{figure}
\centering
\includegraphics[width=3.0in,height=8.0in]{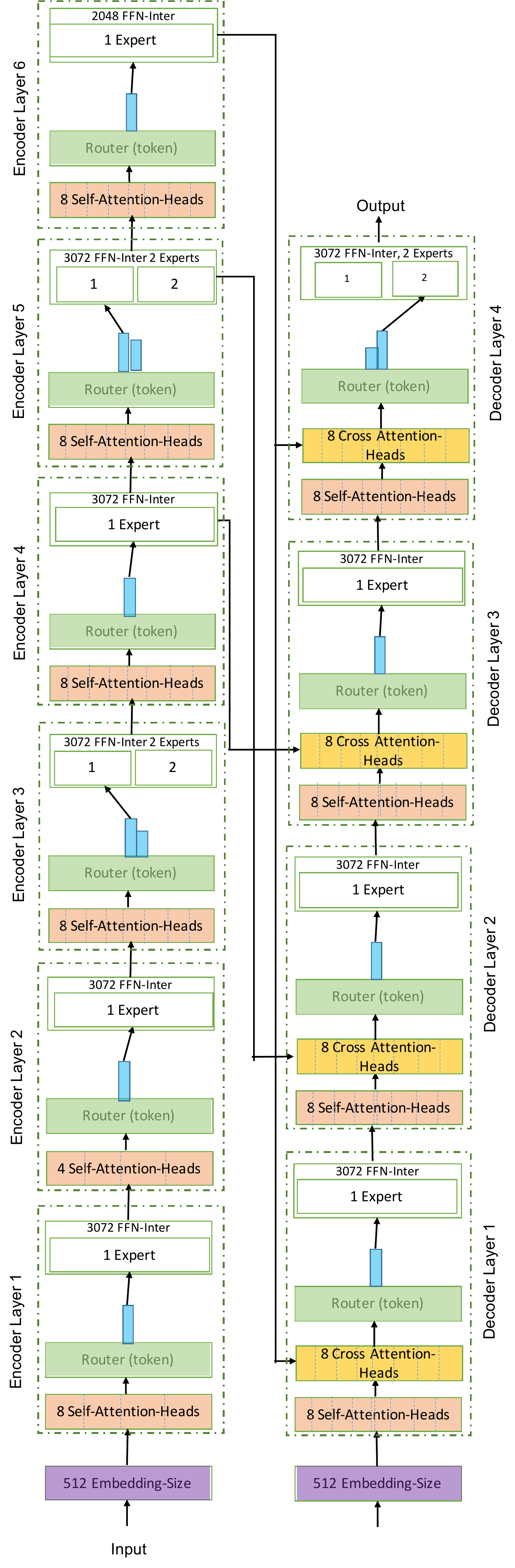}
\vspace{-0.8em}
\caption{{\sysname}-generated architecture for WMT'19 En-De.}
\label{fig:arch_designs_wmt19ende}
\vspace{-2em}
\end{figure}

Figure~\ref{fig:arch_designs_wmt14ende},~\ref{fig:arch_designs_wmt14enfr} and ~\ref{fig:arch_designs_wmt19ende} present the full architecture design of pareto-efficient architectures generated by {\sysname}.

\subsection{Evolutionary Search - Stability}
\label{sec:evosearch_mt_stability}
{We study the initialization effects on the stability of the pareto front outputted by the evolutionary search for HAT. Table~\ref{tab:appd_mt_evosearch_stability} displays sampled (direct) BLEU and latency of the models in the pareto front for different seeds on the WMT'14 En-Fr task. The differences in the latency and BLEU across seeds are mostly marginal. This result highlights that the pareto front outputted by the evolutionary search is largely stable for HAT.}

\begin{table*}
\scriptsize
\begin{center}
\begin{tabular}{cccccccc} \toprule
\textbf{Supernet / Pareto Front} &  &  \multicolumn{2}{c}{\textbf{Model 1}}  & \multicolumn{2}{c}{\textbf{Model 2}} & \multicolumn{2}{c}{\textbf{Model 3}} \\ 
 & \textbf{Seed} & \textbf{Latency} & \textbf{BLEU} & \textbf{Latency} & \textbf{BLEU} & \textbf{Latency} & \textbf{BLEU}   \\ 
\midrule
HAT (SPOS) & 1 & 96.39 & 38.94 & 176.44 & 39.26 & 187.53 & 39.16 \\
HAT (SPOS) & 2 & 98.91 & 38.96 & 159.87 & 39.20 & 192.11 & 39.09 \\
HAT (SPOS) & 3 & 100.15 & 38.96 & 158.67 & 39.24 & 189.53 & 39.16 \\
\bottomrule
\end{tabular}
\caption{Stability of the evolutionary search w.r.t. different seeds on the WMT'14 En-Fr task. Search quality is measured in terms of latency and sampled (direct) supernet performance (BLEU) of the models in the pareto front.}
\label{tab:appd_mt_evosearch_stability}
\end{center}
\end{table*}

\subsection{Evolutionary Search - Algorithm}
\label{sec:evosearch_algorithm}
{We present the pseudo code of the evolutionary search algorithm proposed by HAT in Algorithm~\ref{algo:evosearch}. This algorithm is also adopted by AutoMoE.}

\begin{algorithm*}
\hspace*{\algorithmicindent} \textbf{Input:} $\texttt{\mbox{supernet}}$, $\texttt{\mbox{latency-predictor}}$, $\texttt{\mbox{num-iterations}}$, $\texttt{\mbox{num-population}}$, $\texttt{\mbox{num-parents}}$, $\texttt{\mbox{num-mutations}}$, $\texttt{\mbox{num-crossover}}$, $\texttt{\mbox{mutate-prob}}$, $\texttt{\mbox{latency-constraint}}$ \\
     \hspace*{\algorithmicindent} \textbf{Output:} $\texttt{\mbox{best-architecture}}$ 
     \begin{algorithmic}[1]
     \State $popu \gets$ $\texttt{\mbox{num-population}}$ random samples from the search space
     \For{$iter \gets 1 \: \texttt{\mbox{to}} \: \texttt{\mbox{num-iterations}}$}
        \State $\texttt{\mbox{cur-parents}} \gets$ top `$\texttt{\mbox{num-parents}}$' architectures from $popu$ by supernet's validation loss
        \State $\texttt{\mbox{cur-mutate-popu}}$ = $\{ \}$
        \For{$mi \gets 1 \: \texttt{\mbox{to}} \: \texttt{\mbox{num-mutations}}$}
            \State $\texttt{\mbox{cur-mutate-gene}} \gets$ mutate a random example from $popu$ with mutation probability $\texttt{\mbox{mutate-prob}}$
            \If{$\texttt{\mbox{cur-mutate-gene}}$ satisfies $\texttt{\mbox{latency-constraint}}$ via $\texttt{\mbox{latency-predictor}}$}
               \State $\texttt{\mbox{cur-mutate-popu}} = \texttt{\mbox{cur-mutate-popu}} \cup \texttt{\mbox{cur-mutate-gene}}$
            \EndIf
        \EndFor
        \State $\texttt{\mbox{cur-crossover-popu}}$ = $\{ \}$
        \For{$ci \gets 1 \: \texttt{\mbox{to}} \: \texttt{\mbox{num-crossover}}$}
            \State $\texttt{\mbox{cur-crossover-gene}} \gets$ crossover two random examples from $popu$
            \If{$\texttt{\mbox{cur-crossover-gene}}$ satisfies $\texttt{\mbox{latency-constraint}}$ via $\texttt{\mbox{latency-predictor}}$}
               \State $\texttt{\mbox{cur-crossover-popu}} = \texttt{\mbox{cur-crossover-popu}} \cup \texttt{\mbox{cur-crossover-gene}}$
            \EndIf
        \EndFor
        \State $popu = \texttt{\mbox{cur-parents}} \cup \texttt{\mbox{cur-mutate-popu}} \cup \texttt{\mbox{cur-crossover-popu}}$
     \EndFor
     \State return top architecture from $popu$ by supernet's validation loss
\end{algorithmic}
\caption{Evolutionary search algorithm for Neural architecture search.}
\label{algo:evosearch}
\end{algorithm*}

%% file: main.bbl
\begin{thebibliography}{39}
\expandafter\ifx\csname natexlab\endcsname\relax\def\natexlab#1{#1}\fi

\bibitem[{Artetxe et~al.(2021)Artetxe, Bhosale, Goyal, Mihaylov, Ott, Shleifer,
  Lin, Du, Iyer, Pasunuru, Anantharaman, Li, Chen, Akin, Baines, Martin, Zhou,
  Koura, O'Horo, Wang, Zettlemoyer, Diab, Kozareva, and
  Stoyanov}]{artetxe_arxiv21}
Mikel Artetxe, Shruti Bhosale, Naman Goyal, Todor Mihaylov, Myle Ott, Sam
  Shleifer, Xi~Victoria Lin, Jingfei Du, Srinivasan Iyer, Ramakanth Pasunuru,
  Giri Anantharaman, Xian Li, Shuohui Chen, Halil Akin, Mandeep Baines, Louis
  Martin, Xing Zhou, Punit~Singh Koura, Brian O'Horo, Jeff Wang, Luke
  Zettlemoyer, Mona~T. Diab, Zornitsa Kozareva, and Ves Stoyanov. 2021.
\newblock \href {http://arxiv.org/abs/2112.10684} {Efficient large scale
  language modeling with mixtures of experts}.
\newblock \emph{CoRR}, abs/2112.10684.

\bibitem[{Brown et~al.(2020)Brown, Mann, Ryder, Subbiah, Kaplan, Dhariwal,
  Neelakantan, Shyam, Sastry, Askell, Agarwal, Herbert-Voss, Krueger, Henighan,
  Child, Ramesh, Ziegler, Wu, Winter, Hesse, Chen, Sigler, Litwin, Gray, Chess,
  Clark, Berner, McCandlish, Radford, Sutskever, and Amodei}]{gpt3}
Tom Brown, Benjamin Mann, Nick Ryder, Melanie Subbiah, Jared~D Kaplan, Prafulla
  Dhariwal, Arvind Neelakantan, Pranav Shyam, Girish Sastry, Amanda Askell,
  Sandhini Agarwal, Ariel Herbert-Voss, Gretchen Krueger, Tom Henighan, Rewon
  Child, Aditya Ramesh, Daniel Ziegler, Jeffrey Wu, Clemens Winter, Chris
  Hesse, Mark Chen, Eric Sigler, Mateusz Litwin, Scott Gray, Benjamin Chess,
  Jack Clark, Christopher Berner, Sam McCandlish, Alec Radford, Ilya Sutskever,
  and Dario Amodei. 2020.
\newblock \href
  {https://proceedings.neurips.cc/paper/2020/file/1457c0d6bfcb4967418bfb8ac142f64a-Paper.pdf}
  {Language models are few-shot learners}.
\newblock In \emph{Advances in Neural Information Processing Systems},
  volume~33, pages 1877--1901. Curran Associates, Inc.

\bibitem[{Cai et~al.(2020)Cai, Gan, Wang, Zhang, and Han}]{onceforall}
Han Cai, Chuang Gan, Tianzhe Wang, Zhekai Zhang, and Song Han. 2020.
\newblock \href {https://arxiv.org/pdf/1908.09791.pdf} {Once for all: Train one
  network and specialize it for efficient deployment}.
\newblock In \emph{International Conference on Learning Representations}.

\bibitem[{Devlin et~al.(2019)Devlin, Chang, Lee, and Toutanova}]{bert}
Jacob Devlin, Ming-Wei Chang, Kenton Lee, and Kristina Toutanova. 2019.
\newblock \href {https://doi.org/10.18653/v1/N19-1423} {{BERT}: Pre-training of
  deep bidirectional transformers for language understanding}.
\newblock In \emph{Proceedings of the 2019 Conference of the North {A}merican
  Chapter of the Association for Computational Linguistics: Human Language
  Technologies, Volume 1 (Long and Short Papers)}, pages 4171--4186,
  Minneapolis, Minnesota. Association for Computational Linguistics.

\bibitem[{Dong et~al.(2021)Dong, Wang, Xu, Peng, Ren, and
  Liang}]{efficient_bert}
Chenhe Dong, Guangrun Wang, Hang Xu, Jiefeng Peng, Xiaozhe Ren, and Xiaodan
  Liang. 2021.
\newblock \href {https://doi.org/10.18653/v1/2021.findings-emnlp.123}
  {{E}fficient{BERT}: Progressively searching multilayer perceptron via warm-up
  knowledge distillation}.
\newblock In \emph{Findings of the Association for Computational Linguistics:
  EMNLP 2021}, pages 1424--1437, Punta Cana, Dominican Republic. Association
  for Computational Linguistics.

\bibitem[{Du et~al.(2022)Du, Huang, Dai, Tong, Lepikhin, Xu, Krikun, Zhou, Yu,
  Firat, Zoph, Fedus, Bosma, Zhou, Wang, Wang, Webster, Pellat, Robinson,
  Meier-Hellstern, Duke, Dixon, Zhang, Le, Wu, Chen, and Cui}]{glam}
Nan Du, Yanping Huang, Andrew~M Dai, Simon Tong, Dmitry Lepikhin, Yuanzhong Xu,
  Maxim Krikun, Yanqi Zhou, Adams~Wei Yu, Orhan Firat, Barret Zoph, Liam Fedus,
  Maarten~P Bosma, Zongwei Zhou, Tao Wang, Emma Wang, Kellie Webster, Marie
  Pellat, Kevin Robinson, Kathleen Meier-Hellstern, Toju Duke, Lucas Dixon, Kun
  Zhang, Quoc Le, Yonghui Wu, Zhifeng Chen, and Claire Cui. 2022.
\newblock \href {https://proceedings.mlr.press/v162/du22c.html} {{GL}a{M}:
  Efficient scaling of language models with mixture-of-experts}.
\newblock In \emph{Proceedings of the 39th International Conference on Machine
  Learning}, volume 162 of \emph{Proceedings of Machine Learning Research},
  pages 5547--5569. PMLR.

\bibitem[{Fedus et~al.(2022{\natexlab{a}})Fedus, Dean, and
  Zoph}]{fedus_survey_arxiv22}
William Fedus, Jeff Dean, and Barret Zoph. 2022{\natexlab{a}}.
\newblock \href {https://doi.org/10.48550/ARXIV.2209.01667} {A review of sparse
  expert models in deep learning}.

\bibitem[{Fedus et~al.(2022{\natexlab{b}})Fedus, Zoph, and
  Shazeer}]{switchtransformers}
William Fedus, Barret Zoph, and Noam Shazeer. 2022{\natexlab{b}}.
\newblock \href {http://jmlr.org/papers/v23/21-0998.html} {Switch transformers:
  Scaling to trillion parameter models with simple and efficient sparsity}.
\newblock \emph{Journal of Machine Learning Research}, 23(120):1--39.

\bibitem[{Gao et~al.(2022)Gao, Xu, Shi, Ren, Yu, Liang, Jiang, and
  Li}]{autobert_zero}
Jiahui Gao, Hang Xu, Han Shi, Xiaozhe Ren, Philip L.~H. Yu, Xiaodan Liang, Xin
  Jiang, and Zhenguo Li. 2022.
\newblock \href {https://ojs.aaai.org/index.php/AAAI/article/view/21311}
  {Autobert-zero: Evolving {BERT} backbone from scratch}.
\newblock In \emph{Thirty-Sixth {AAAI} Conference on Artificial Intelligence,
  {AAAI} 2022, Thirty-Fourth Conference on Innovative Applications of
  Artificial Intelligence, {IAAI} 2022, The Twelveth Symposium on Educational
  Advances in Artificial Intelligence, {EAAI} 2022 Virtual Event, February 22 -
  March 1, 2022}, pages 10663--10671. {AAAI} Press.

\bibitem[{Gong et~al.(2022)Gong, Wang, Li, Chen, Yan, Tian, qiang liu, and
  Chandra}]{nasvit}
Chengyue Gong, Dilin Wang, Meng Li, Xinlei Chen, Zhicheng Yan, Yuandong Tian,
  qiang liu, and Vikas Chandra. 2022.
\newblock \href {https://openreview.net/forum?id=Qaw16njk6L} {{NASV}it: Neural
  architecture search for efficient vision transformers with gradient conflict
  aware supernet training}.
\newblock In \emph{International Conference on Learning Representations}.

\bibitem[{Guo et~al.(2020)Guo, Zhang, Mu, Heng, Liu, Wei, and Sun}]{spos}
Zichao Guo, Xiangyu Zhang, Haoyuan Mu, Wen Heng, Zechun Liu, Yichen Wei, and
  Jian Sun. 2020.
\newblock \href {https://doi.org/10.1007/978-3-030-58517-4\_32} {Single path
  one-shot neural architecture search with uniform sampling}.
\newblock In \emph{Computer Vision - {ECCV} 2020 - 16th European Conference,
  Glasgow, UK, August 23-28, 2020, Proceedings, Part {XVI}}, volume 12361 of
  \emph{Lecture Notes in Computer Science}, pages 544--560. Springer.

\bibitem[{Hua et~al.(2022)Hua, Dai, Liu, and Le}]{flashattention}
Weizhe Hua, Zihang Dai, Hanxiao Liu, and Quoc Le. 2022.
\newblock Transformer quality in linear time.
\newblock In \emph{Proceedings of the 39th International Conference on Machine
  Learning}, volume 162 of \emph{Proceedings of Machine Learning Research},
  pages 9099--9117. PMLR.

\bibitem[{Javaheripi et~al.(2022)Javaheripi, Shah, Mukherjee, Religa, Mendes,
  de~Rosa, Bubeck, Koushanfar, and Dey}]{litetransformersearch}
Mojan Javaheripi, Shital Shah, Subhabrata Mukherjee, Tomasz~L. Religa, Caio
  C.~T. Mendes, Gustavo~H. de~Rosa, Sebastien Bubeck, Farinaz Koushanfar, and
  Debadeepta Dey. 2022.
\newblock \href {https://doi.org/10.48550/ARXIV.2203.02094}
  {Litetransformersearch: Training-free on-device search for efficient
  autoregressive language models}.

\bibitem[{Kasai et~al.(2021)Kasai, Pappas, Peng, Cross, and
  Smith}]{kasai2021deep}
Jungo Kasai, Nikolaos Pappas, Hao Peng, James Cross, and Noah Smith. 2021.
\newblock \href {https://openreview.net/forum?id=KpfasTaLUpq} {Deep encoder,
  shallow decoder: Reevaluating non-autoregressive machine translation}.
\newblock In \emph{International Conference on Learning Representations}.

\bibitem[{Kim et~al.(2021)Kim, Awan, Muzio, Cruz{-}Salinas, Lu, Hendy,
  Rajbhandari, He, and Awadalla}]{kim_etal_21}
Young~Jin Kim, Ammar~Ahmad Awan, Alexandre Muzio, Andr{\'{e}}s~Felipe
  Cruz{-}Salinas, Liyang Lu, Amr Hendy, Samyam Rajbhandari, Yuxiong He, and
  Hany~Hassan Awadalla. 2021.
\newblock \href {http://arxiv.org/abs/2109.10465} {Scalable and efficient moe
  training for multitask multilingual models}.
\newblock \emph{CoRR}, abs/2109.10465.

\bibitem[{Kudugunta et~al.(2021)Kudugunta, Huang, Bapna, Krikun, Lepikhin,
  Luong, and Firat}]{kudugunta_etal_21}
Sneha Kudugunta, Yanping Huang, Ankur Bapna, Maxim Krikun, Dmitry Lepikhin,
  Minh-Thang Luong, and Orhan Firat. 2021.
\newblock \href {https://doi.org/10.18653/v1/2021.findings-emnlp.304} {Beyond
  distillation: Task-level mixture-of-experts for efficient inference}.
\newblock In \emph{Findings of the Association for Computational Linguistics:
  EMNLP 2021}, pages 3577--3599, Punta Cana, Dominican Republic. Association
  for Computational Linguistics.

\bibitem[{Lepikhin et~al.(2020)Lepikhin, Lee, Xu, Chen, Firat, Huang, Krikun,
  Shazeer, and Chen}]{gshard}
Dmitry Lepikhin, HyoukJoong Lee, Yuanzhong Xu, Dehao Chen, Orhan Firat, Yanping
  Huang, Maxim Krikun, Noam Shazeer, and Zhifeng Chen. 2020.
\newblock \href {http://arxiv.org/abs/2006.16668} {Gshard: Scaling giant models
  with conditional computation and automatic sharding}.
\newblock \emph{CoRR}, abs/2006.16668.

\bibitem[{Lewis et~al.(2021)Lewis, Bhosale, Dettmers, Goyal, and
  Zettlemoyer}]{base}
Mike Lewis, Shruti Bhosale, Tim Dettmers, Naman Goyal, and Luke Zettlemoyer.
  2021.
\newblock \href {https://proceedings.mlr.press/v139/lewis21a.html} {Base
  layers: Simplifying training of large, sparse models}.
\newblock In \emph{Proceedings of the 38th International Conference on Machine
  Learning}, volume 139 of \emph{Proceedings of Machine Learning Research},
  pages 6265--6274. PMLR.

\bibitem[{Liu et~al.(2021)Liu, Dai, So, and Le}]{gmlp}
Hanxiao Liu, Zihang Dai, David So, and Quoc~V Le. 2021.
\newblock \href
  {https://proceedings.neurips.cc/paper/2021/file/4cc05b35c2f937c5bd9e7d41d3686fff-Paper.pdf}
  {Pay attention to mlps}.
\newblock In \emph{Advances in Neural Information Processing Systems},
  volume~34, pages 9204--9215. Curran Associates, Inc.

\bibitem[{Liu et~al.(2022)Liu, Kim, Muzio, and Hassan}]{gating_dropout}
Rui Liu, Young~Jin Kim, Alexandre Muzio, and Hany Hassan. 2022.
\newblock \href {https://proceedings.mlr.press/v162/liu22g.html} {Gating
  dropout: Communication-efficient regularization for sparsely activated
  transformers}.
\newblock In \emph{Proceedings of the 39th International Conference on Machine
  Learning}, volume 162 of \emph{Proceedings of Machine Learning Research},
  pages 13782--13792. PMLR.

\bibitem[{Ott et~al.(2019)Ott, Edunov, Baevski, Fan, Gross, Ng, Grangier, and
  Auli}]{fairseq}
Myle Ott, Sergey Edunov, Alexei Baevski, Angela Fan, Sam Gross, Nathan Ng,
  David Grangier, and Michael Auli. 2019.
\newblock \href {https://doi.org/10.18653/v1/N19-4009} {fairseq: A fast,
  extensible toolkit for sequence modeling}.
\newblock In \emph{Proceedings of the 2019 Conference of the North {A}merican
  Chapter of the Association for Computational Linguistics (Demonstrations)},
  pages 48--53, Minneapolis, Minnesota. Association for Computational
  Linguistics.

\bibitem[{Papineni et~al.(2002)Papineni, Roukos, Ward, and Zhu}]{bleuscore}
Kishore Papineni, Salim Roukos, Todd Ward, and Wei-Jing Zhu. 2002.
\newblock \href {https://doi.org/10.3115/1073083.1073135} {{B}leu: a method for
  automatic evaluation of machine translation}.
\newblock In \emph{Proceedings of the 40th Annual Meeting of the Association
  for Computational Linguistics}, pages 311--318, Philadelphia, Pennsylvania,
  USA. Association for Computational Linguistics.

\bibitem[{Rajbhandari et~al.(2022)Rajbhandari, Li, Yao, Zhang, Aminabadi, Awan,
  Rasley, and He}]{deepspeed_moe}
Samyam Rajbhandari, Conglong Li, Zhewei Yao, Minjia Zhang, Reza~Yazdani
  Aminabadi, Ammar~Ahmad Awan, Jeff Rasley, and Yuxiong He. 2022.
\newblock \href {https://proceedings.mlr.press/v162/rajbhandari22a.html}
  {{D}eep{S}peed-{M}o{E}: Advancing mixture-of-experts inference and training
  to power next-generation {AI} scale}.
\newblock In \emph{Proceedings of the 39th International Conference on Machine
  Learning}, volume 162 of \emph{Proceedings of Machine Learning Research},
  pages 18332--18346. PMLR.

\bibitem[{Roller et~al.(2021)Roller, Sukhbaatar, Szlam, and Weston}]{hash}
Stephen Roller, Sainbayar Sukhbaatar, Arthur Szlam, and Jason~E Weston. 2021.
\newblock \href {https://openreview.net/forum?id=lMgDDWb1ULW} {Hash layers for
  large sparse models}.
\newblock In \emph{Advances in Neural Information Processing Systems}.

\bibitem[{Schuster et~al.(2022)Schuster, Fisch, Gupta, Dehghani, Bahri, Tran,
  Tay, and Metzler}]{calm}
Tal Schuster, Adam Fisch, Jai Gupta, Mostafa Dehghani, Dara Bahri, Vinh~Q.
  Tran, Yi~Tay, and Donald Metzler. 2022.
\newblock \href {https://doi.org/10.48550/ARXIV.2207.07061} {Confident adaptive
  language modeling}.

\bibitem[{Shazeer et~al.(2017)Shazeer, Mirhoseini, Maziarz, Davis, Le, Hinton,
  and Dean}]{shazeer2017}
Noam Shazeer, *Azalia Mirhoseini, *Krzysztof Maziarz, Andy Davis, Quoc Le,
  Geoffrey Hinton, and Jeff Dean. 2017.
\newblock \href {https://openreview.net/forum?id=B1ckMDqlg} {Outrageously large
  neural networks: The sparsely-gated mixture-of-experts layer}.
\newblock In \emph{International Conference on Learning Representations}.

\bibitem[{So et~al.(2019)So, Le, and Liang}]{evotrans}
David So, Quoc Le, and Chen Liang. 2019.
\newblock \href {https://proceedings.mlr.press/v97/so19a.html} {The evolved
  transformer}.
\newblock In \emph{Proceedings of the 36th International Conference on Machine
  Learning}, volume~97 of \emph{Proceedings of Machine Learning Research},
  pages 5877--5886. PMLR.

\bibitem[{So et~al.(2021)So, Ma\'{n}ke, Liu, Dai, Shazeer, and Le}]{primer}
David So, Wojciech Ma\'{n}ke, Hanxiao Liu, Zihang Dai, Noam Shazeer, and Quoc~V
  Le. 2021.
\newblock \href
  {https://proceedings.neurips.cc/paper/2021/file/2f3c6a4cd8af177f6456e7e51a916ff3-Paper.pdf}
  {Searching for efficient transformers for language modeling}.
\newblock In \emph{Advances in Neural Information Processing Systems},
  volume~34, pages 6010--6022. Curran Associates, Inc.

\bibitem[{Vaswani et~al.(2017)Vaswani, Shazeer, Parmar, Uszkoreit, Jones,
  Gomez, Kaiser, and Polosukhin}]{NIPS2017_3f5ee243}
Ashish Vaswani, Noam Shazeer, Niki Parmar, Jakob Uszkoreit, Llion Jones,
  Aidan~N Gomez, \L~ukasz Kaiser, and Illia Polosukhin. 2017.
\newblock \href
  {https://proceedings.neurips.cc/paper/2017/file/3f5ee243547dee91fbd053c1c4a845aa-Paper.pdf}
  {Attention is all you need}.
\newblock In \emph{Advances in Neural Information Processing Systems},
  volume~30. Curran Associates, Inc.

\bibitem[{Wang et~al.(2018)Wang, Singh, Michael, Hill, Levy, and
  Bowman}]{wang-etal-2018-glue}
Alex Wang, Amanpreet Singh, Julian Michael, Felix Hill, Omer Levy, and Samuel
  Bowman. 2018.
\newblock \href {https://doi.org/10.18653/v1/W18-5446} {{GLUE}: A multi-task
  benchmark and analysis platform for natural language understanding}.
\newblock In \emph{Proceedings of the 2018 {EMNLP} Workshop {B}lackbox{NLP}:
  Analyzing and Interpreting Neural Networks for {NLP}}, pages 353--355,
  Brussels, Belgium. Association for Computational Linguistics.

\bibitem[{Wang et~al.(2020)Wang, Wu, Liu, Cai, Zhu, Gan, and Han}]{hat}
Hanrui Wang, Zhanghao Wu, Zhijian Liu, Han Cai, Ligeng Zhu, Chuang Gan, and
  Song Han. 2020.
\newblock \href {https://doi.org/10.18653/v1/2020.acl-main.686} {{HAT}:
  Hardware-aware transformers for efficient natural language processing}.
\newblock In \emph{Proceedings of the 58th Annual Meeting of the Association
  for Computational Linguistics}, pages 7675--7688, Online. Association for
  Computational Linguistics.

\bibitem[{Xu et~al.(2022{\natexlab{a}})Xu, Mukherjee, Liu, Dey, Wang, Zhang,
  Awadallah, and Gao}]{autodistill}
Dongkuan Xu, Subhabrata~(Subho) Mukherjee, Xiaodong Liu, Debadeepta Dey, Wenhui
  Wang, Xiang Zhang, Ahmed~H. Awadallah, and Jianfeng Gao. 2022{\natexlab{a}}.
\newblock \href {https://arxiv.org/abs/2201.08539} {Autodistil: Few-shot
  task-agnostic neural architecture search for distilling large language
  models}.
\newblock \emph{ArXiv}.

\bibitem[{Xu et~al.(2021)Xu, Tan, Luo, Song, Li, Qin, and Liu}]{nasbert}
Jin Xu, Xu~Tan, Renqian Luo, Kaitao Song, Jian Li, Tao Qin, and Tie-Yan Liu.
  2021.
\newblock \href {https://doi.org/10.1145/3447548.3467262} {Nas-bert:
  Task-agnostic and adaptive-size bert compression with neural architecture
  search}.
\newblock In \emph{Proceedings of the 27th ACM SIGKDD Conference on Knowledge
  Discovery \& Data Mining}, KDD '21, page 1933–1943, New York, NY, USA.
  Association for Computing Machinery.

\bibitem[{Xu et~al.(2022{\natexlab{b}})Xu, Tan, Song, Luo, Leng, Qin, Liu, and
  Li}]{magic}
Jin Xu, Xu~Tan, Kaitao Song, Renqian Luo, Yichong Leng, Tao Qin, Tie-Yan Liu,
  and Jian Li. 2022{\natexlab{b}}.
\newblock \href {https://proceedings.mlr.press/v162/xu22h.html} {Analyzing and
  mitigating interference in neural architecture search}.
\newblock In \emph{Proceedings of the 39th International Conference on Machine
  Learning}, volume 162 of \emph{Proceedings of Machine Learning Research},
  pages 24646--24662. PMLR.

\bibitem[{Yin et~al.(2021)Yin, Chen, Shang, Jiang, Chen, and
  Liu}]{autotinybert}
Yichun Yin, Cheng Chen, Lifeng Shang, Xin Jiang, Xiao Chen, and Qun Liu. 2021.
\newblock \href {https://doi.org/10.18653/v1/2021.acl-long.400}
  {{A}uto{T}iny{BERT}: Automatic hyper-parameter optimization for efficient
  pre-trained language models}.
\newblock In \emph{Proceedings of the 59th Annual Meeting of the Association
  for Computational Linguistics and the 11th International Joint Conference on
  Natural Language Processing (Volume 1: Long Papers)}, pages 5146--5157,
  Online. Association for Computational Linguistics.

\bibitem[{Yu et~al.(2019)Yu, Yang, Xu, Yang, and Huang}]{sandwich}
Jiahui Yu, Linjie Yang, Ning Xu, Jianchao Yang, and Thomas Huang. 2019.
\newblock \href {https://openreview.net/forum?id=H1gMCsAqY7} {Slimmable neural
  networks}.
\newblock In \emph{International Conference on Learning Representations}.

\bibitem[{Yuksel et~al.(2012)Yuksel, Wilson, and Gader}]{20yearsmoe}
Seniha~Esen Yuksel, Joseph~N. Wilson, and Paul~D. Gader. 2012.
\newblock \href {https://doi.org/10.1109/TNNLS.2012.2200299} {Twenty years of
  mixture of experts}.
\newblock \emph{IEEE Transactions on Neural Networks and Learning Systems},
  23(8):1177--1193.

\bibitem[{Zoph et~al.(2022)Zoph, Bello, Kumar, Du, Huang, Dean, Shazeer, and
  Fedus}]{stmoe}
Barret Zoph, Irwan Bello, Sameer Kumar, Nan Du, Yanping Huang, Jeff Dean, Noam
  Shazeer, and William Fedus. 2022.
\newblock \href {https://doi.org/10.48550/ARXIV.2202.08906} {St-moe: Designing
  stable and transferable sparse expert models}.

\bibitem[{Zuo et~al.(2022)Zuo, Liu, Jiao, Kim, Hassan, Zhang, Gao, and
  Zhao}]{stochasticexperts}
Simiao Zuo, Xiaodong Liu, Jian Jiao, Young~Jin Kim, Hany Hassan, Ruofei Zhang,
  Jianfeng Gao, and Tuo Zhao. 2022.
\newblock \href {https://openreview.net/forum?id=B72HXs80q4} {Taming sparsely
  activated transformer with stochastic experts}.
\newblock In \emph{International Conference on Learning Representations}.

\end{thebibliography}
